\documentclass{article}

\usepackage[final]{corl_2026}

\usepackage{graphicx}
\usepackage{amsmath}
\usepackage{amssymb}
\usepackage{booktabs}
\usepackage{multirow}
\usepackage{wrapfig}
\usepackage{subcaption}
\usepackage{algorithm}
\usepackage{algorithmic}
\usepackage{enumitem}
\usepackage{needspace}
\usepackage{xcolor}

\definecolor{crvblue}{RGB}{0,72,180}
\newcommand{\crv}[1]{{#1}}
\makeatletter
\renewcommand{\@conferencelocation}{Austin, Texas, USA}
\makeatother
\hypersetup{colorlinks=true,linkcolor=black,citecolor=black,urlcolor=black,
  pdftitle={Beyond Imitation: Reinforcement Learning-Based Sim-Real Co-Training for VLA Models},
  pdfauthor={Liangzhi Shi, Shuaihang Chen, Feng Gao, Yinuo Chen, Kang Chen, Tonghe Zhang, Hongzhi Zang, Jiakai Zhou, Weinan Zhang, Chao Yu, Yu Wang}}

\setlist[itemize]{leftmargin=1.3em,labelsep=0.45em,itemsep=0pt,topsep=2pt}

\newcommand{\modify}[1]{#1}

\title{%
\raisebox{-0.16\height}[0pt][0pt]{\includegraphics[width=0.102\linewidth]{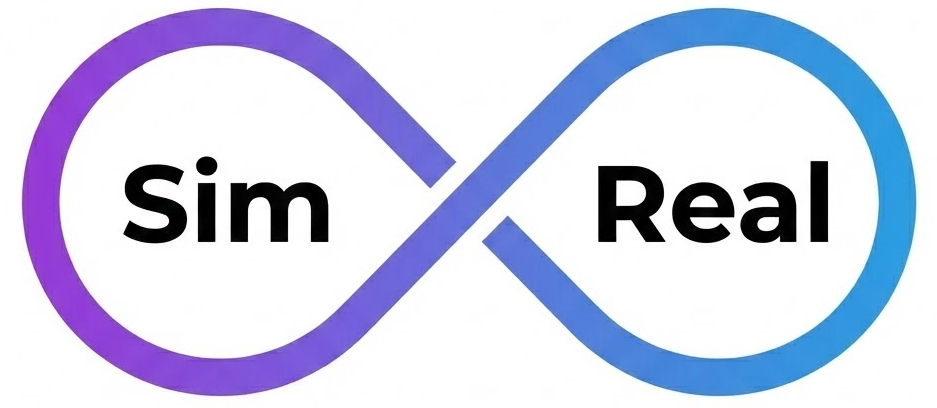}}\hspace{0.3em}Beyond Imitation:\\
Reinforcement Learning-Based Sim-Real\\
Co-Training for VLA Models}

\author{
\crv{\textbf{Liangzhi Shi$^{1,5,*}$, Shuaihang Chen$^{2,6,*}$, Feng Gao$^{1}$, Yinuo Chen$^{1}$,}}\\
\crv{\textbf{Kang Chen$^{3,6}$, Tonghe Zhang$^{4}$, Hongzhi Zang$^{1}$, Jiakai Zhou$^{1}$,}}\\
\crv{\textbf{Weinan Zhang$^{2}$, Chao Yu$^{1,\dagger}$, Yu Wang$^{1}$}}\\[1em]
\crv{$^{1}$Tsinghua University \quad $^{2}$Harbin Institute of Technology \quad $^{3}$Peking University}\\
\crv{$^{4}$Carnegie Mellon University \quad $^{5}$Shanghai AI Laboratory \quad $^{6}$Zhongguancun Academy}\\[1em]
\crv{$^*$Equal Contribution \quad $^\dagger$Corresponding author: \texttt{yuchao@sz.tsinghua.edu.cn}}\\[0.7em]
{\hypersetup{urlcolor=black}\color{black}Project Page: \href{https://rl-co-training.github.io/}{\nolinkurl{https://rl-co-training.github.io/}}}
}

\begin{document}
\maketitle

\begin{abstract}
Simulation offers a scalable and low-cost way to enrich vision-language-action (VLA) training, reducing reliance on expensive real-robot demonstrations. However, most sim-real co-training methods rely on supervised fine-tuning (SFT), which treats simulation as a static source of demonstrations and does not exploit large-scale closed-loop interaction. Consequently, real-world gains and generalization are often limited. In this paper, we propose an \underline{\textit{RL}}-based sim-real \underline{\textit{Co}}-training \modify{(RL-Co)} framework that leverages interactive simulation while preserving real-world capabilities. Our method follows a generic two-stage design: we first warm-start the policy with SFT on a mixture of real and simulated demonstrations, then fine-tune it with reinforcement learning in simulation while adding an auxiliary supervised loss on real-world data to anchor the policy and mitigate catastrophic forgetting. We evaluate our framework on four real-world tabletop manipulation tasks using two representative VLA architectures, OpenVLA and $\pi_{0.5}$, and observe consistent improvements over real-only fine-tuning and SFT-based co-training, including +24\% real-world success on OpenVLA and +20\% on $\pi_{0.5}$. Beyond higher success rates, RL co-training yields stronger generalization to unseen task variations
and substantially improved real-world data efficiency, providing a practical and scalable pathway for leveraging simulation to enhance real-robot deployment.

\end{abstract}


\section{Introduction}
\label{sec:intro}

Vision-language-action (VLA) models are a promising route toward general-purpose robots, integrating visual observations, language instructions, and action generation in a unified policy framework~\cite{brohan2022rt, zitkovich2023rt, kim2024openvla, black2024pi_0, intelligence2025pi_, team2024octo}.
Large-scale robot demonstrations and visual-language priors support efficient supervised fine-tuning~\cite{o2024open, khazatsky2024droid, walke2023bridgedata, kim2025fine}, yet policies adapted from limited real demonstrations remain brittle under novel scenes, object variations, and closed-loop execution shifts~\cite{zhang2025vlabench, liu2025can}.
Collecting more real data can help, but remains slow, expensive, and difficult to scale~\cite{o2024open, khazatsky2024droid}.

Simulation can address this bottleneck by cheaply scaling environments, trajectories, and closed-loop interaction~\cite{todorov2012mujoco, makoviychuk2021isaac, tao2024maniskill3}, but simulation-trained policies still face the sim-to-real gap.
Domain randomization improves robustness by varying visual and physical factors~\cite{tobin2017domain, peng2018sim, andrychowicz2020learning}, while real-to-sim-to-real and reconstruction-based pipelines improve simulator fidelity through scene reconstruction, system identification, and digital twins~\cite{torne2024reconcilingrealitysimulationrealtosimtoreal, wu2025rl, zhang2025real, li2024robogsim}.
These approaches reduce the gap, but often require carefully designed randomization, accurate reconstruction, or task-specific simulator tuning.

Sim-real co-training offers a different route by introducing real-world data into training instead of requiring direct transfer from a simulation-trained policy.
Prior work aligns sim-real representations or joint observation-action distributions~\cite{cheng2025generalizable, yang2025invariance, yu2024natural, lei2026mechanistic}, and complementary studies show that even low-fidelity or weakly aligned simulation can improve real-world policies when it expands the task-relevant training distribution~\cite{maddukuri2025sim, wei2025empirical}.
This makes co-training a scalable alternative to high-fidelity sim-to-real pipelines, but most existing methods still treat simulation as a static source of demonstrations, leaving closed-loop interaction underexplored.

\begin{wrapfigure}{r}{0.5\linewidth}

    \centering
    \includegraphics[width=0.8\linewidth]{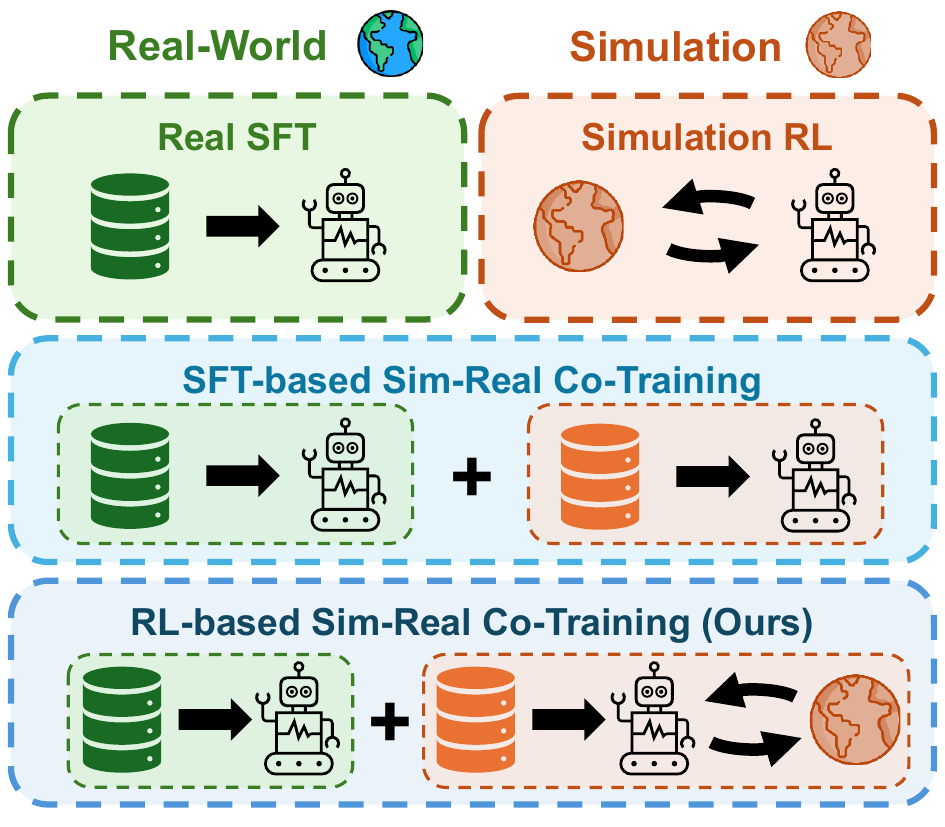}
    \caption{
    \textbf{Training paradigms.}
    \modify{RL-Co} uses simulation for RL interaction while anchoring updates with real-world SFT.
    }
    \label{fig:intro}

\end{wrapfigure}

Interactive policy improvement directly targets this limitation.
Reinforcement learning can reduce the closed-loop distribution shift of pure imitation through trial-and-error interaction~\cite{ross2011reduction, osa2018algorithmic}, but robotic manipulation often involves sparse rewards, long horizons, and costly exploration~\cite{dulac2019challenges}.
Combining imitation with RL, either through demonstration initialization or behavior-prior regularization, is therefore a common remedy~\cite{zhu2018reinforcement, rajeswaran2018learning, fujimoto2021minimalist, lu2023imitation, li2025qchunking, zhang2025rewind}.
Recent VLA and large robot policy post-training methods similarly use supervised fine-tuning before RL improves task success and generalization~\cite{liu2025can, li2025simplevla, zhang2025reinflow, zhang2025sac, zang2026rlinfvlaunifiedefficientframework}.
However, these methods usually improve the policy within one interaction domain and, when applied to sim-to-real settings, still rely on zero-shot transfer.
Sim-real VLA post-training must instead exploit cheap simulation interaction while preventing drift from real visual observations, action distributions, and execution priors.

In this work, we propose an RL-based sim-real co-training framework for VLA models that turns simulation into an interactive post-training source while using real demonstrations as cross-domain anchors.
The policy is first initialized through supervised sim-real co-training, then improved by reinforcement learning in simulation with an auxiliary real-world SFT objective that keeps simulation-driven updates grounded in deployable observations and actions.

We evaluate RL-Co on real-world tabletop manipulation tasks with OpenVLA~\cite{kim2024openvla} and $\pi_{0.5}$~\cite{intelligence2025pi_}.
Across tasks and model families, RL-Co improves real-world success over real-only SFT and SFT-based sim-real co-training \crv{using simulation demonstrations generated by MimicGen}, while also improving generalization to unseen task variations, reducing real-demonstration needs, and providing gains beyond visual diversity from heavy domain randomization or data augmentation.

\section{Related Works}

\subsection{Vision-Language-Action Models for Manipulation}

Recent vision-language-action (VLA) models formulate manipulation as conditional action generation from visual observations and language instructions~\cite{brohan2022rt, zitkovich2023rt, kim2024openvla, black2024pi_0, intelligence2025pi_, team2024octo}. Their progress relies on large-scale robot demonstrations and visual-language priors~\cite{o2024open, khazatsky2024droid, walke2023bridgedata}, but deployment often still requires task-specific supervised fine-tuning~\cite{kim2025fine} and remains sensitive to shifts across scenes, objects, and execution states~\cite{zhang2025vlabench, liu2025can}. These limitations motivate post-training methods that improve robustness and generalization beyond static supervised data.

\subsection{Interactive and Regularized Post-Training for Robot Policies}

Supervised fine-tuning or behavior cloning aligns policies with target demonstrations, but it is vulnerable to covariate shift and compounding closed-loop errors~\cite{ross2011reduction, osa2018algorithmic}. Reinforcement learning provides an interactive alternative and has long been used for robot policy learning~\cite{kalashnikov2018qtoptscalabledeepreinforcement, zang2026rlinfuser}. \crv{SERL~\cite{luo2024serl} demonstrates efficient real-robot learning, while DSRL~\cite{crv_dsrl} steers diffusion-policy noise, RL-100~\cite{crv_rl100} combines offline and online RL for diffusion policies, $\pi^{*}_{0.6}$~\cite{amin2025pi06} learns from deployment experience and expert corrections, and RLT~\cite{crv_rltoken} trains a lightweight actor--critic on VLA representations.} \crv{However,} direct real-world RL is often constrained by safety, reset cost, and sample efficiency~\cite{dulac2019challenges}.

Recent work therefore moves RL post-training for larger robot policies and VLA models into simulation\crv{. Empirical studies demonstrate gains in VLA generalization~\cite{liu2025can}, while FLaRe~\cite{hu2025flare} and SimpleVLA-RL~\cite{li2025simplevla} scale RL fine-tuning of pretrained robot policies. For flow-based policies, ReinFlow~\cite{zhang2025reinflow} enables likelihood-based online updates, SAC Flow~\cite{zhang2025sac} stabilizes off-policy learning, and $\pi_{\texttt{RL}}$~\cite{chen2026pitextttrlonlinerlfinetuning} develops online RL fine-tuning for flow-based VLAs.}

In parallel, offline, online, and robot-adaptation methods show that reward-driven updates can be stabilized by imitation-style losses, behavior constraints, action priors, or demonstration-derived reward and policy priors~\cite{rajeswaran2018learning, fujimoto2021minimalist, lu2023imitation, li2025qchunking, zhang2025rewind}, while residual policy learning offers another way to refine imitation-learned manipulation behaviors~\cite{ankile2025imitation}. \crv{PLD~\cite{crv_pld} trains residual RL specialists around a frozen VLA and uses their collected trajectories for supervised fine-tuning of the VLA.} These works establish the value of combining interaction-driven improvement with behavior priors, but they typically operate in a single deployment domain or rely on zero-shot sim-to-real transfer. RL-Co instead uses simulation for interaction while anchoring VLA updates with real demonstrations for physical deployment.

\subsection{Sim-to-Real Transfer and Sim-Real Co-Training}

Simulation provides a safe and scalable platform for robot learning, and several sim-to-real RL methods use it to accelerate real-world adaptation~\cite{eysenbach2021offdynamics, yin2025sgft, daoudi2024conservative, lin2025simtoreal}. \crv{Direct transfer can succeed with suitable simulation diversity and training scale~\cite{crv_omnireset,crv_molmobot}, although visual and dynamics discrepancies remain challenges.} Domain randomization addresses this gap through broad visual or physical variations~\cite{tobin2017domain, peng2018sim, andrychowicz2020learning, chebotar2019closing, mehta2020active}, while real-to-sim-to-real and reconstruction-based pipelines improve simulator fidelity through scene reconstruction, system identification, or digital twins~\cite{torne2024reconcilingrealitysimulationrealtosimtoreal, wu2025rl, zhang2025real, li2024robogsim}; both can require careful randomization, environment construction, or tuning.

Beyond direct transfer, sim-real co-training introduces real-world data by learning domain-invariant or task-relevant representations~\cite{cheng2025generalizable, yang2025invariance, yu2024natural, lei2026mechanistic}, or by treating simulation as a scalable source of demonstrations and trajectory variation\crv{~\cite{mandlekar2023mimicgen,crv_skillgen,jiang2024dexmimicen,crv_pgdg}} that can improve real-world policies despite limited visual fidelity or task correspondence~\cite{maddukuri2025sim, wei2025empirical, chen2025robotwin, nasiriany2024robocasa}.

These approaches show that imperfect simulation can provide useful training data, but mostly exploit it through supervised data mixing rather than closed-loop policy improvement. RL-Co builds on this data-augmentation view while incorporating reinforcement learning into co-training, enabling active exploration in simulation and grounding the policy with real-world data.

\section{Preliminaries}

\subsection{Problem Setup and VLA Post-Training}
\label{sec:problem-vla-post-training}

We study a two-domain manipulation setting with a real-world task domain $T_{\text{real}}$ and a corresponding simulation domain $T_{\text{sim}}$.
The two task-level counterparts share the robot embodiment, action interface, language instruction, and task-level initial-state distribution, while allowing visual and dynamics discrepancies.
We model both as partially observable decision processes.
A vision-language-action (VLA) policy $\pi_{\theta}$ maps recent observations and the language instruction to robot actions.

We use two standard VLA post-training primitives: supervised fine-tuning (SFT) on task demonstrations~\cite{kim2024openvla, kim2025fine, black2024pi_0} and reinforcement learning (RL) through reward-guided interaction.
\modify{RL-Co} combines them so that SFT provides initialization and real-domain anchoring, while RL supplies closed-loop improvement in simulation.
More detailed problem definitions and post-training objectives are provided in Appendix~\ref{app:problem-formulation}.

\begin{figure*}
    \centering
    \includegraphics[width=\linewidth]{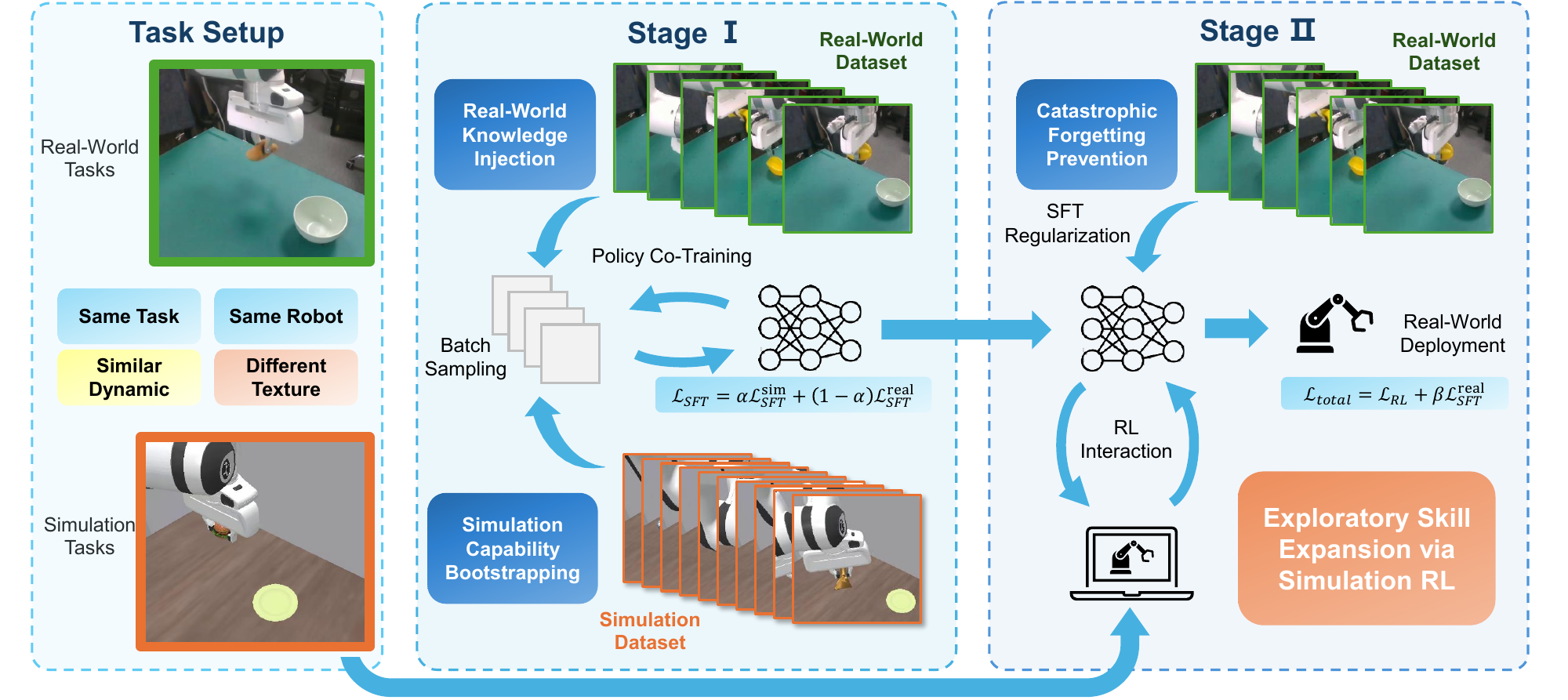}
    \caption{\textbf{Overview of RL-Co.}
    Stage~I initializes the VLA policy with mixed real and simulated demonstrations.
    Stage~II fine-tunes it with simulation RL while using real-world SFT as a regularizer to preserve deployable behaviors.}
    \label{fig:overview}
\end{figure*}

\subsection{SFT-based Co-Training}
\label{sec:sft-co-training}

Given a real-world dataset $\mathcal{D}_{\text{real}}$ and a simulation dataset $\mathcal{D}_{\text{sim}}$, SFT-based sim-real co-training jointly fine-tunes the VLA policy on both sources of demonstrations:
\begin{equation}
\mathcal{L}_{\text{SFT}}(\theta)
=
\alpha \, \mathcal{L}_{\text{SFT}}(\theta; \mathcal{D}_{\text{sim}})
+
(1-\alpha) \, \mathcal{L}_{\text{SFT}}(\theta; \mathcal{D}_{\text{real}}),
\end{equation}
where $\alpha \in [0,1]$ controls the relative contribution of simulated data.
Following \citet{maddukuri2025sim}, this objective can be implemented by sampling training trajectories from $\mathcal{D}_{\text{sim}}$ with probability $\alpha$ and from $\mathcal{D}_{\text{real}}$ with probability $1-\alpha$.

This SFT-based co-training strategy is a strong baseline for leveraging simulation as additional supervised data.
However, because it remains an imitation objective, it cannot explicitly use reward feedback or online interaction, motivating the reinforcement learning-based co-training approach introduced next.

\section{Method}

We present our \modify{\underline{\textit{RL}}-based sim-real \underline{\textit{Co}}-training (RL-Co)} framework, as summarized in Fig.~\ref{fig:overview}.
RL-Co follows a two-stage design.
In Stage~I, we initialize the policy through supervised co-training on both real-world and simulated demonstrations.
In Stage~II, we further improve the policy with reinforcement learning in simulation, while explicitly preserving real-world capabilities through an auxiliary supervised objective.

\subsection{Stage I: SFT Co-Training for Policy Initialization}

Starting from a pre-trained VLA policy $\pi_{\theta}$ that has not been adapted to the target tasks, Stage~I initializes the policy with both real-world and simulated demonstrations.
Specifically, we apply the SFT co-training objective in Section~\ref{sec:sft-co-training} to the real-world dataset $\mathcal{D}_{\text{real}}$ and the simulation dataset $\mathcal{D}_{\text{sim}}$.
This stage serves two purposes: it lets the policy rapidly absorb task-specific real-world knowledge needed for deployment, and gives the policy sufficient simulation competence for non-trivial RL initialization by learning from simulated demonstrations.
These properties motivate SFT co-training as the first stage of RL-Co; Section~\ref{sec:ablation-study} provides a detailed analysis of its contribution.

\subsection{Stage II: \modify{Sim-Real Co-Training with Real-Regularized RL}}

While Stage~I equips the policy with both real-world and simulated capabilities, its optimization remains limited to imitation.
Stage~II therefore seeks to expand the policy's competence through online interaction in simulation, while preventing degradation of real-world performance.
To achieve this, we introduce an auxiliary supervised fine-tuning objective on real-world data into the RL fine-tuning process.
During simulation training, each policy update is driven by a reinforcement learning loss $\mathcal{L}_{\text{RL}}$, which encourages exploration and maximizes task rewards.
We augment this objective with an SFT loss on $\mathcal{D}_{\text{real}}$, yielding
\begin{equation}
\mathcal{L}_{\text{total}}
=
\mathcal{L}_{\text{RL}}
+
\beta \, \mathcal{L}_{\text{SFT}}(\theta; \mathcal{D}_{\text{real}}),
\end{equation}
where $\beta$ balances reinforcement learning updates and preservation of real-world knowledge.

Intuitively, the RL term lets the policy exploit large-scale simulated interaction to explore diverse behaviors and improve task performance, while the real-world supervision term regularizes the update by anchoring the policy to real demonstrations and mitigating catastrophic forgetting.
This simple yet effective modification is compatible with a wide range of RL fine-tuning algorithms and forms the core of \modify{RL-Co}.

\begin{figure}[t]
    \centering
    \begin{subfigure}[t]{0.44\linewidth}
        \centering
        \includegraphics[width=\linewidth]{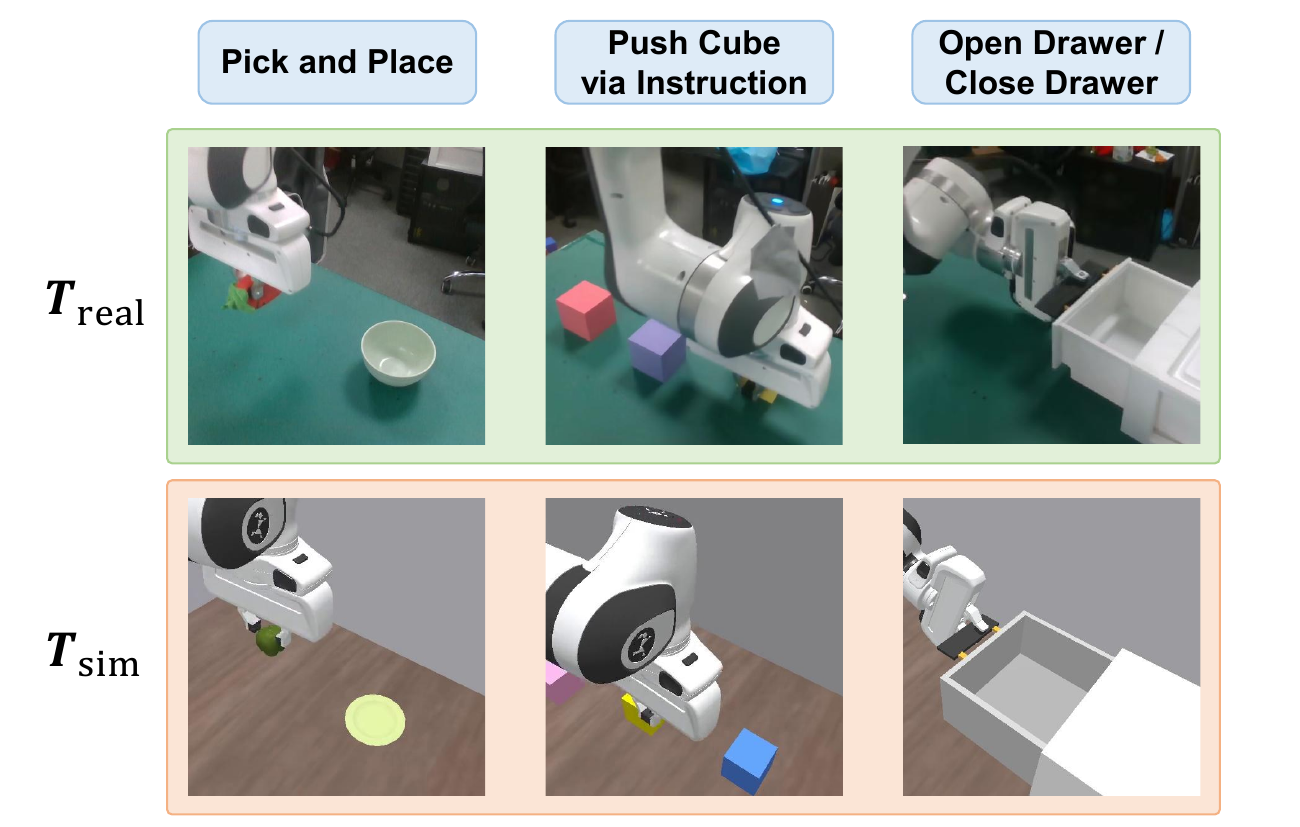}
        \caption{Sim-real task setup.}
        \label{fig:env_sim_real}
    \end{subfigure}
    \hfill
    \begin{subfigure}[t]{0.52\linewidth}
        \centering
        \includegraphics[width=\linewidth]{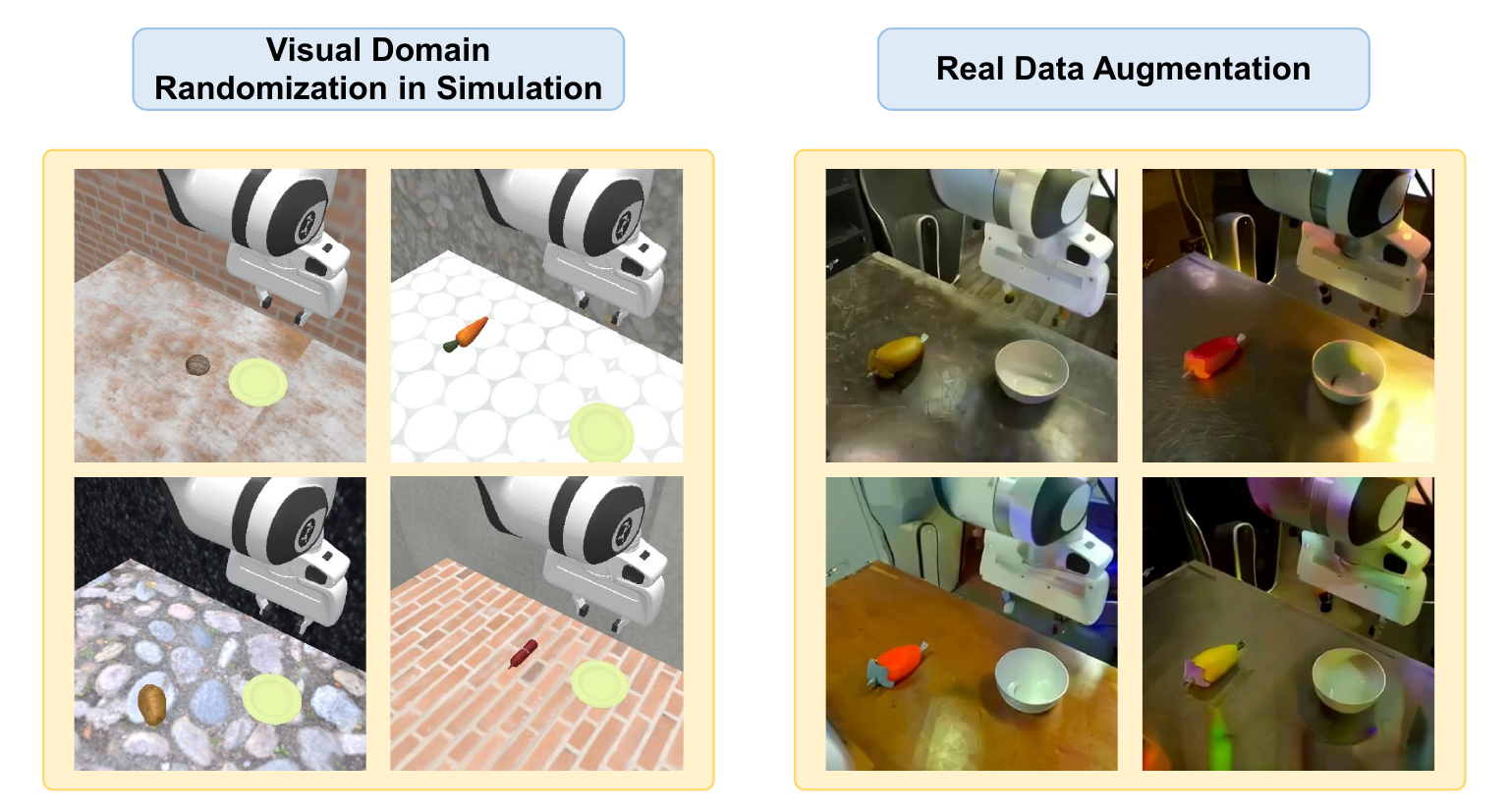}
        \caption{Visual diversity.}
        \label{fig:env_visual_diversity}
    \end{subfigure}
    \caption{\textbf{Environment visualizations.}
    (a) Real and simulated tabletop tasks used in our sim-real co-training experiments.
    (b) Examples of visual domain-randomized simulation and data-augmented real observations used to test whether visual diversity alone explains the gains.}
    \label{fig:env}

\end{figure}

\section{Experiments}


\crv{In this section, we empirically evaluate the proposed RL-Co framework and ask:}
\begin{itemize}[leftmargin=1.05em,labelwidth=0.55em,labelsep=0.5em,labelindent=0pt,align=left,itemsep=3pt,topsep=4pt,parsep=0pt]
    \item \crv{Does RL-Co improve real-world effectiveness and robustness across tasks and VLA architectures?}
    \item \crv{Why is the two-stage design necessary for transferring simulation interaction to real-world deployment?}
    \item \crv{To what extent can RL-Co reduce the amount of required real-world demonstration data?}
\end{itemize}

We answer these questions in order: First, section~\ref{sec:real-world-effectiveness} compares \modify{RL-Co} with real-only SFT and SFT-based sim--real co-training, evaluates robustness under unseen settings, and tests whether stronger visual diversity alone explains the gains. Next, section~\ref{sec:ablation-study} analyzes the key components of \modify{RL-Co}, including simulation SFT initialization, real-world supervision, and the hyperparameters $\alpha$ and $\beta$. Finally, section~\ref{sec:real-data} studies data efficiency by varying the number of real-world demonstrations.

\subsection{Experimental Setting}

\textbf{Environmental setting.}
We evaluate \modify{RL-Co} on four tabletop manipulation tasks that require perception, language grounding, and control: \texttt{Pick and Place}, \texttt{Push Cube via Instruction}, \texttt{Open Drawer}, and \texttt{Close Drawer}.
Fig.~\ref{fig:env}(a) visualizes the real and simulated environments.
Both domains use a Franka Emika Panda robot with 7-DoF end-effector delta control; real-world observations are captured by a fixed RGB camera, and all methods are evaluated on the same independently sampled initial states with task success rate as the metric.
We build the simulation environments in ManiSkill~\cite{tao2024maniskill3} and align them with the real setup at the task level, while allowing differences in object identity, materials, textures, and lighting.
Detailed real-world setup, task definitions, evaluation protocol, and simulation configuration are provided in Appendices~\ref{app:real-world-setup}--\ref{app:simulation-training}.

\crv{\textbf{Evaluation protocol.}
We evaluate one trained checkpoint per condition across two sessions. Each session contains 32 trials for \texttt{Pick and Place}, 30 for \texttt{Push Cube}, 10 for \texttt{Open Drawer}, and 10 for \texttt{Close Drawer}. Appendix~\ref{app:crv-evaluation} details the aggregation of session-level results. The real setup was fixed before constructing and aligning the simulator.}

\textbf{Data generation.}
For each real-world task, we collect 20--50 successful expert demonstrations via human teleoperation, forming $\mathcal{D}_{\text{real}}$.
In simulation, we use MimicGen~\cite{mandlekar2023mimicgen} with replayed real-world trajectories as seeds to generate 1{,}000 successful trajectories per task, forming $\mathcal{D}_{\text{sim}}$.
Further details on real trajectory collection, seed replay, and trajectory pruning are provided in Appendix~\ref{app:dataset-generation}.

\textbf{Implementation.}
We instantiate \modify{RL-Co} on two representative VLA families: next-token prediction-based OpenVLA~\cite{kim2024openvla} and flow-matching-based $\pi_{0.5}$~\cite{intelligence2025pi_}.
In Stage~I, we directly mix real-world and simulation datasets and train with the official SFT implementations of each model.
In Stage~II, OpenVLA follows the \crv{PPO~\cite{schulman2017proximal}} fine-tuning protocol and implementation of~\citet{liu2025can} with \crv{LoRA~\cite{hu2022lora} and} our real-world regularization loss, whereas $\pi_{0.5}$ uses ReinFlow~\cite{zhang2025reinflow} implemented with RLinf~\cite{yu2025rlinf}, \crv{updating only its action expert}.
We train each model-task pair with a fixed interaction budget and run simulation RL to convergence.

\subsection{Real-World Effectiveness and Robustness}
\label{sec:real-world-effectiveness}

\subsubsection{Main Comparison Across Tasks and VLA Models}

We first evaluate whether \modify{RL-Co} improves real-world deployment under limited real demonstrations and imperfect simulation.
\crv{We compare against real-only SFT and SFT co-training with static MimicGen demonstrations. RL-Co uses the same Stage~I data, followed by online simulation rollouts and the real-data SFT anchor in Stage~II.}
Table~\ref{tab:co_training_results} reports the quantitative results.

\begin{table*}
    \centering
    \resizebox{\textwidth}{!}{%
    \begin{tabular}{c|c|cccc|c}
    \toprule
        VLA Model
        & Experiment Setting
        & \texttt{Pick and Place}
        & \texttt{Push Cube}
        & \texttt{Open Drawer}
        & \texttt{Close Drawer}
        & \texttt{Avg} \\
    \midrule
        \multirow{3}{*}{OpenVLA}
        & Real-Only Training
        & 6.3 $\pm$ 0.0  & 20.0 $\pm$ 13.3 & 0.0 $\pm$ 0.0 & 10.0 $\pm$ 10.0 & \crv{9.1 $\pm$ 5.8} \\
        & SFT Co-Training
        & 23.4 $\pm$ 4.7 & 51.7 $\pm$ 5.0 & 0.0 $\pm$ 0.0 & 85.0 $\pm$ 5.0 & 40.0 $\pm$ 3.7 \\
        & \textbf{\modify{RL-Co} (Ours)}
        & \crv{\textbf{57.8} $\pm$ 10.9} & \textbf{68.3} $\pm$ 11.7 & \textbf{35.0} $\pm$ 15.0 & \textbf{95.0} $\pm$ 5.0 & \textbf{64.0} $\pm$ 0.7 \\
    \midrule
        \multirow{3}{*}{$\pi_{0.5}$}
        & Real-Only Training
        & 71.9 $\pm$ 9.4 & 0.0 $\pm$ 0.0 & 0.0 $\pm$ 0.0 & 35.0 $\pm$ 15.0 & 26.7 $\pm$ 1.4 \\
        & SFT Co-Training
        & 68.8 $\pm$ 9.4 & 10.0 $\pm$ 3.3 & 10.0 $\pm$ 0.0 & 95.0 $\pm$ 5.0 & 45.9 $\pm$ 4.4 \\
        & \textbf{\modify{RL-Co} (Ours)}
        & \textbf{81.3} $\pm$ 9.4 & \crv{\textbf{18.3} $\pm$ 1.7} & \textbf{65.0} $\pm$ 5.0 & \textbf{100.0} $\pm$ 0.0 & \crv{\textbf{66.1} $\pm$ 4.0} \\
    \bottomrule
    \end{tabular}
    }
    \caption{ \textbf{Comparison of real-world success rates under different training paradigms.}
We compare our \modify{RL-Co} approach with real-only SFT and SFT co-training across four tabletop manipulation tasks, evaluated on both OpenVLA and $\pi_{0.5}$. Results are reported in terms of success rate (SR, $\%$). \crv{Values are the mean $\pm$ population standard deviation over two evaluation sessions of one trained checkpoint. Avg macro-averages tasks within each session before computing the mean and standard deviation.}
}
    \label{tab:co_training_results}
\end{table*}

Real-only SFT is weak under limited demonstrations, especially for OpenVLA, which remains at or below $20\%$ success in all environments.
The stronger $\pi_{0.5}$ model performs well on the simpler \texttt{Pick and Place} task but still struggles with instruction-conditioned pushing and contact-rich drawer manipulation.

SFT-based sim--real co-training partially helps, for example on \texttt{Close Drawer}, but its gains are inconsistent and can be limited when real-only SFT is already strong, as in \texttt{Pick and Place}.

In contrast, \modify{RL-Co} achieves the strongest overall real-world success, showing that closed-loop simulation RL expands policy capability beyond static simulated demonstrations while real-world SFT regularization keeps the update deployable. \crv{Appendix~\ref{app:crv-rollout} further compares against SFT on strong RL-generated rollouts on \texttt{Open Drawer}, reducing the confound between the optimization objective and simulation-data quality.}

\Needspace{18\baselineskip}
\subsubsection{Generalization Under Distribution Shifts}

\begin{wrapfigure}[15]{r}{0.46\linewidth}
    \vspace{-6pt}
    \centering
    \includegraphics[width=\linewidth]{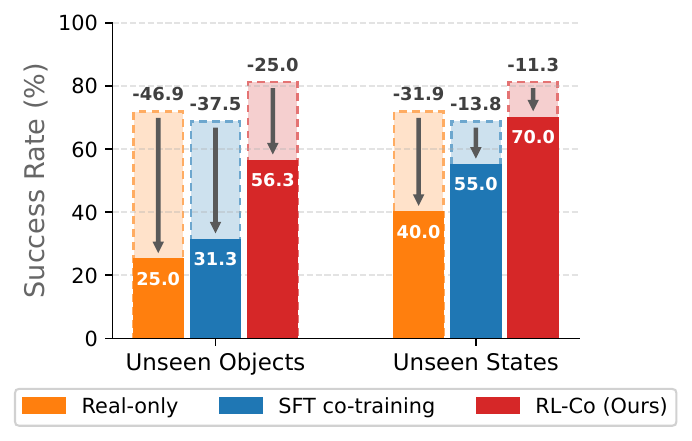}
    \caption{
    \textbf{Comparison of generalization under unseen settings.}
    Solid bars show unseen success rate; pale bars show in-distribution success rate.
    }
    \label{fig:generalization_results}
\end{wrapfigure}

\leavevmode\crv{To further understand how reinforcement learning contributes to improved policy performance, we evaluate generalization under distribution shifts. We focus on \texttt{Pick and Place} with $\pi_{0.5}$, where all three methods achieve comparable in-distribution success, and examine whether their robustness differs under unseen conditions.}
We test \emph{Unseen Objects}, replacing manipulated objects with novel categories, and \emph{Unseen States}, perturbing the robot initial pose beyond the training distribution.
As shown in Fig.~\ref{fig:generalization_results}, real-only SFT degrades sharply under both shifts, while SFT-based sim--real co-training improves robustness by adding simulated variation.
\modify{RL-Co} maintains the strongest unseen performance and the smallest drops, suggesting that interactive simulation helps the policy acquire closed-loop behaviors that transfer more reliably to unseen real-world conditions.

\subsubsection{Targeted Comparison with Stronger Visual-Diversity Baselines}

Having established stronger robustness, we next test whether these gains can be explained by stronger visual diversity alone.
On $\pi_{0.5}$ for \texttt{Pick and Place}, using the same evaluation protocol, we compare against two stronger visual-diversity baselines: heavy visual domain randomization in simulation and video-based augmentation applied to the same real demonstrations.
Fig.~\ref{fig:env}(b) illustrates the visual variations, and Table~\ref{tab:visual_diversity_baselines} reports the results.

\begin{wraptable}{r}{0.5\linewidth}
    \vspace{-6pt}
    \centering
    \small
    \begin{tabular}{lc}
    \toprule
        Method & SR (\%) \\
    \midrule
        Heavy DR zero-shot
        & $10.9 \pm 7.8$ \\
        Cosmos augmentation
        & $67.2 \pm 1.6$ \\
        \textbf{\modify{RL-Co} (Ours)}
        & $\mathbf{81.3 \pm 9.4}$ \\
    \bottomrule
    \end{tabular}
    \caption{
    \textbf{Targeted visual-diversity comparison.}
    Success rates on \texttt{Pick and Place} with $\pi_{0.5}$ under the same real-world evaluation protocol.
    }
    \label{tab:visual_diversity_baselines}

\end{wraptable}

The heavy-DR baseline randomizes the simulated scene with 21 tabletop textures and 21 background textures, independently sampled during training, but still transfers poorly to the real world.
The Cosmos-Transfer~\cite{nvidia2026worldsimulationvideofoundation} augmentation baseline generates 1{,}000 visually augmented trajectories from the original real demonstrations and mixes them with the real data for SFT.
Its performance remains below \modify{RL-Co}, indicating that visual-only augmentation without additional action-level behavior or closed-loop interaction is insufficient. In contrast, \modify{RL-Co} expands the policy's action-level behavioral repertoire through reward-driven interaction in simulation, thereby improving the model's task capability beyond what can be achieved by visual augmentation alone.

\subsection{Mechanistic Analysis and Design Ablations}
\label{sec:ablation-study}

\crv{We conduct ablation studies to understand how the components of RL-Co contribute to effective sim--real co-training. We examine the role of simulation data in Stage~I, the contribution of real-world supervision in both stages, and the sensitivity to the co-training ratio and regularization weight.}

\Needspace{5\baselineskip}
\subsubsection{Effect of Simulation SFT Initialization}

\begin{figure}[t]
    \centering
    \begin{minipage}[t]{0.34\linewidth}
        \vspace{0pt}
        \subcaptionbox{Simulation SFT initialization.\label{fig:r_sr}}[\linewidth]{
            \includegraphics[width=\linewidth]{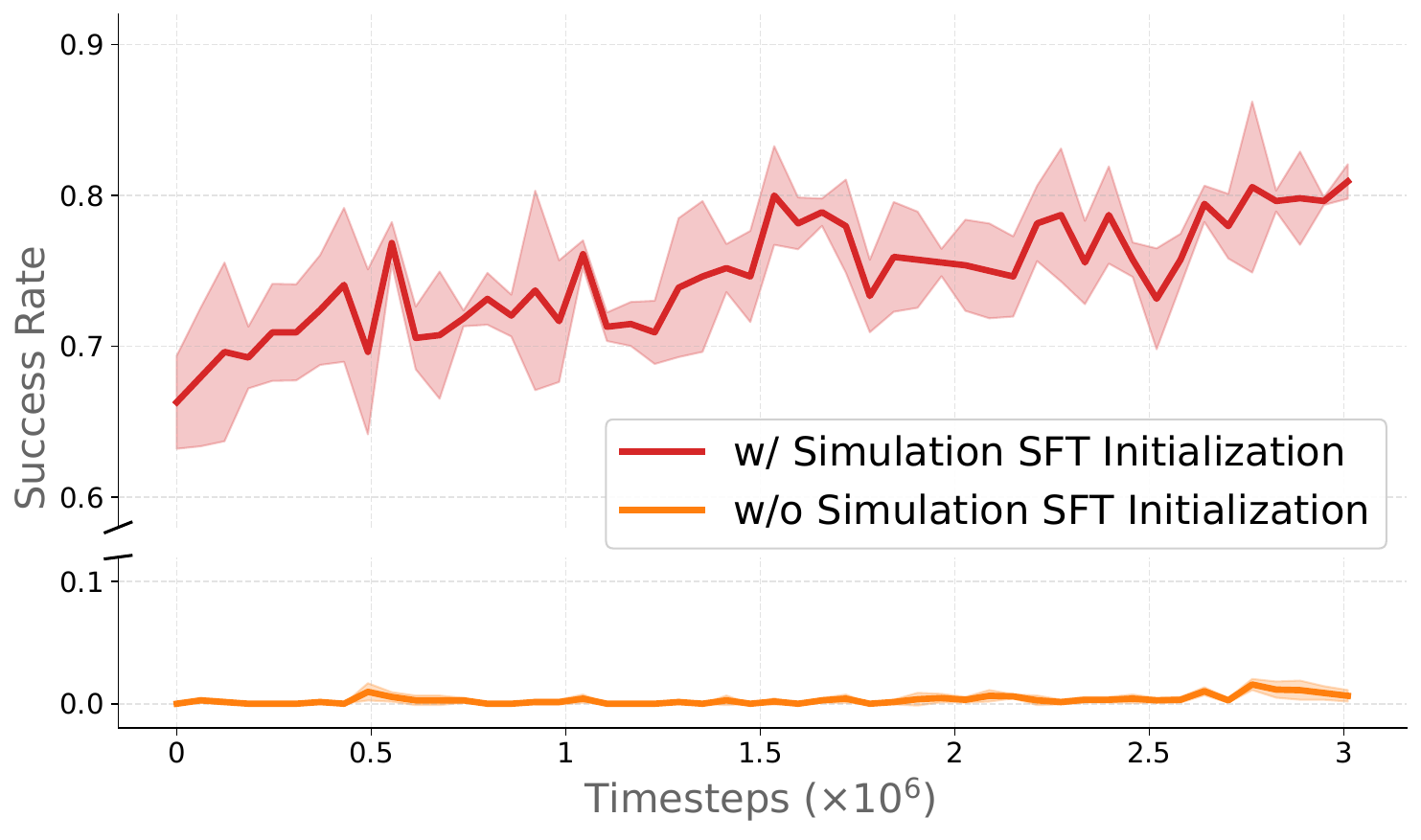}
        }

        \vspace{6pt}
        \subcaptionbox{Real-world supervision.\label{fig:ablation}}[\linewidth]{
            \includegraphics[width=\linewidth]{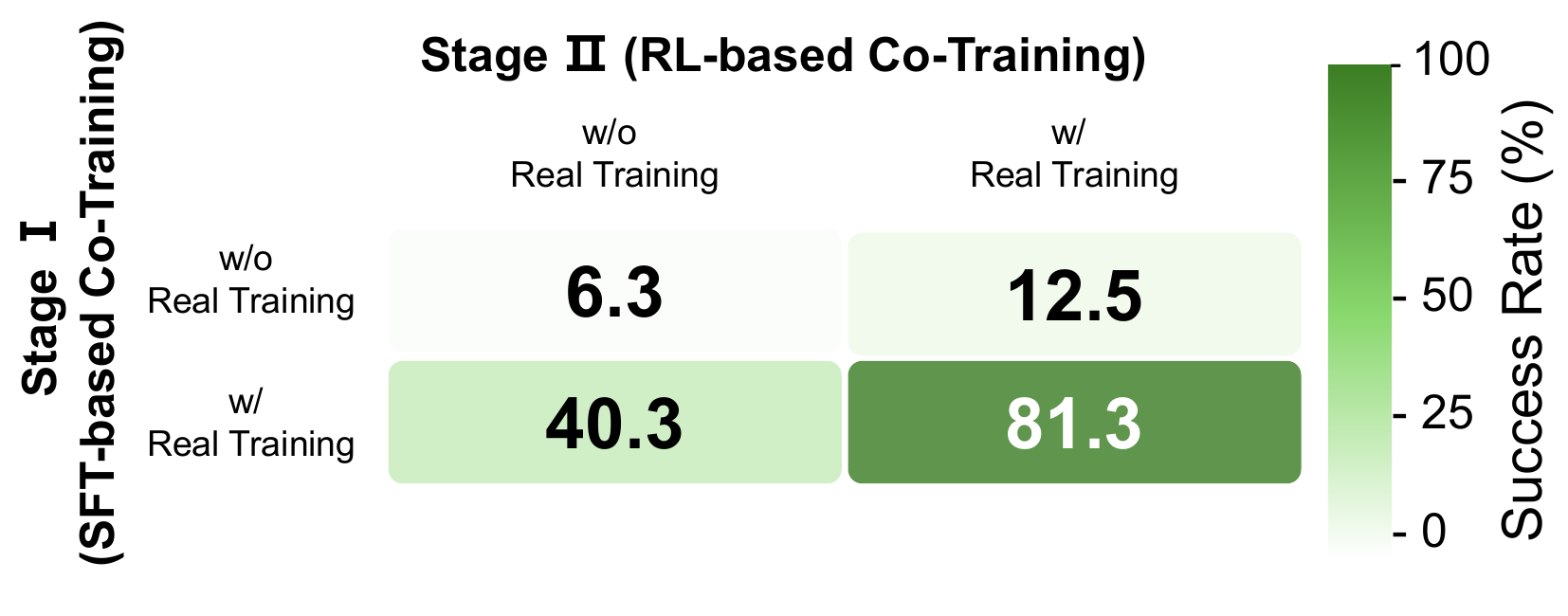}
        }
    \end{minipage}
    \hfill
    \begin{minipage}[t]{0.62\linewidth}
        \vspace{0pt}
        \subcaptionbox{Co-training ratio $\alpha$ and regularization weight $\beta$.\label{fig:stability}}[\linewidth]{
            \includegraphics[width=\linewidth]{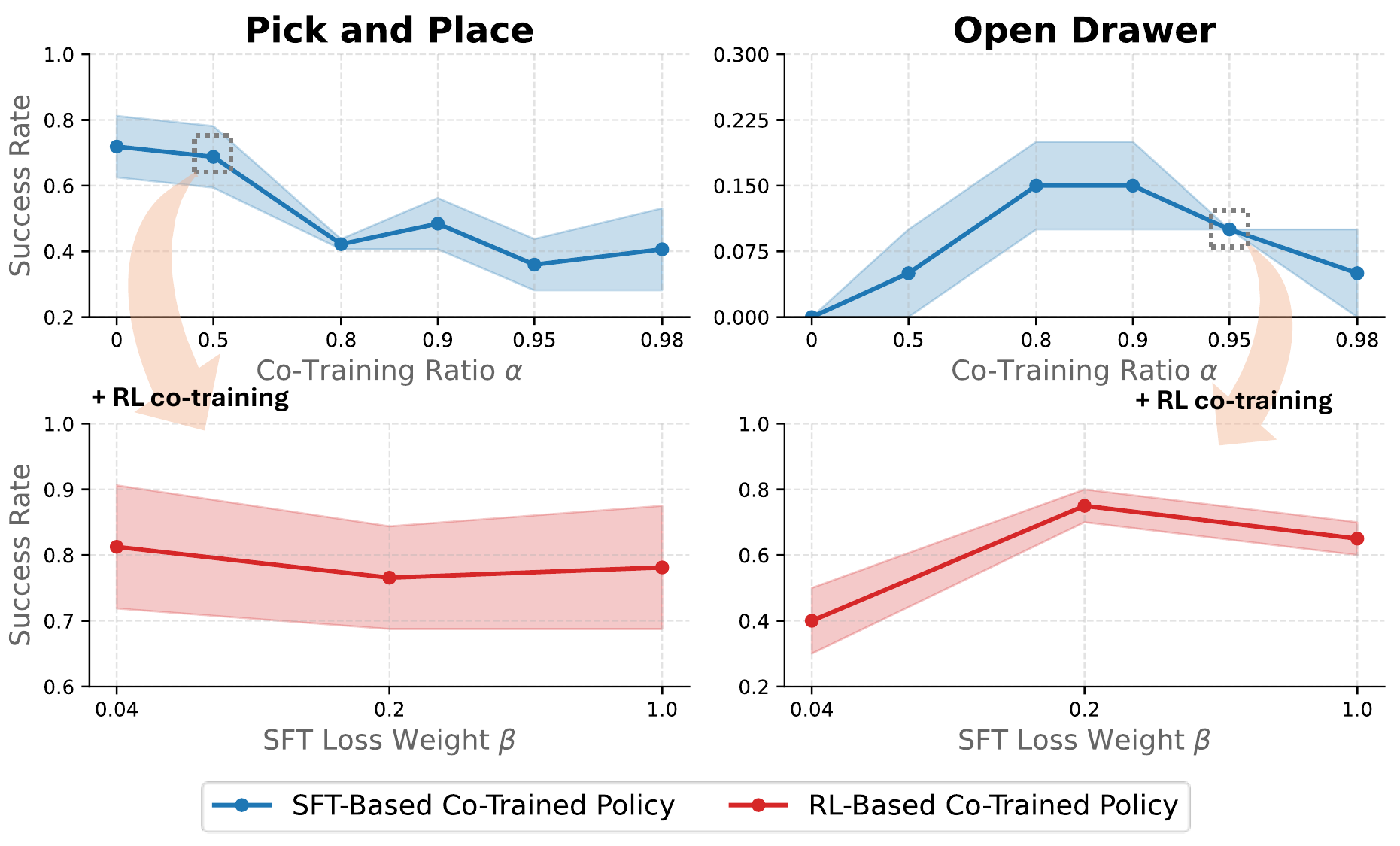}
        }
    \end{minipage}
    \caption{
    \textbf{Mechanistic ablations of \modify{RL-Co}.}
    (a) Simulation SFT initialization enables efficient RL training.
    (b) Real-world supervision in both stages improves deployment.
    (c) RL co-training remains effective across co-training ratios $\alpha$ and regularization weights $\beta$.
    Shaded regions indicate standard deviation where applicable.
    }
    \label{fig:ablation_summary}

\end{figure}

\crv{To evaluate the necessity of simulation data in Stage~I, we perform RL co-training starting from a policy trained only with real-world demonstrations.}
Fig.~\ref{fig:r_sr} compares RL co-training from real-only SFT initialization against full sim--real SFT initialization.

Without simulated SFT initialization, the policy has extremely poor sample efficiency and remains near-trivial after over three million interaction steps.
In contrast, sim--real SFT co-training provides a stronger initialization and enables efficient RL optimization, showing that Stage~I is not merely a data-mixing warm-up but a necessary step for making simulation RL learnable.

\subsubsection{Role of Real-World Supervision in Two Stages}

\crv{Having examined simulation-based initialization, we next assess the role of real-world supervision by removing it from Stage~I and Stage~II respectively.}
Fig.~\ref{fig:ablation} reports real-world success on \texttt{Pick and Place} with $\pi_{0.5}$ under all ablations.
First, removing Stage~II real-world SFT regularization substantially reduces success, indicating domain drift during simulation RL without explicit real-data anchoring.
Removing Stage~I real-world SFT further degrades deployment, suggesting that supervised learning extracts useful priors from limited real demonstrations more efficiently than RL.
\crv{Notably, when real-world supervision is removed from both stages, the policy is trained solely in simulation with limited visual fidelity and achieves only $6.3\%$ success when deployed zero-shot in the real world. This highlights the importance of real-world data:} simulation-only learning is insufficient for deployment without real-data adaptation.

\subsubsection{Sensitivity to Co-Training Ratio alpha and Regularization Weight beta}

\crv{Finally, we examine how the balance of simulated and real-world supervision affects performance. On \texttt{Pick and Place} and \texttt{Open Drawer} with $\pi_{0.5}$, we vary the Stage~I mixture ratio $\alpha$, then select an SFT-co-trained model for each task and vary the Stage~II regularization weight $\beta$.}

As shown in Fig.~\ref{fig:stability}, $\alpha$ strongly affects SFT co-training: on \texttt{Pick and Place}, more simulated data can hurt because real-only training is already strong, whereas on the harder \texttt{Open Drawer} task, neither too little nor too much simulation is optimal.
Nevertheless, across the tested $\beta$ values, RL co-training consistently improves over the corresponding SFT-co-trained policies and exceeds SFT-only models trained under different $\alpha$ settings.
These results indicate that once a reasonable SFT initialization is available, reward-driven interaction can expand the performance limit of supervised co-training, while real-world regularization keeps updates less brittle within the tested range.

\Needspace{19\baselineskip}
\subsection{Real-Data Efficiency}
\label{sec:real-data}

\begin{wrapfigure}[13]{r}{0.5\linewidth}
    \vspace{-6pt}
    \centering
    \includegraphics[width=\linewidth]{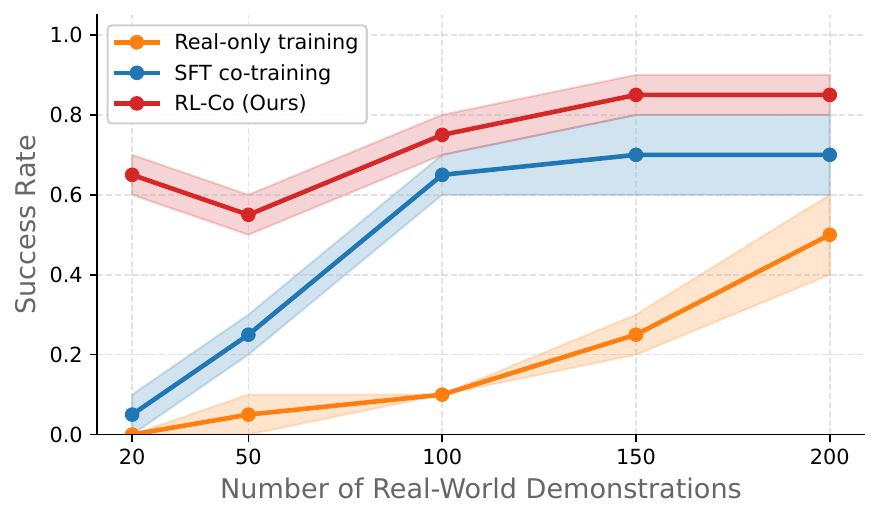}
    \caption{
    \textbf{Real-data efficiency.}
    Shaded regions indicate standard deviation.
    }
    \label{fig:data_eef}

\end{wrapfigure}

We finally study whether \modify{RL-Co} can reduce the amount of real-world demonstration data required for deployment.
Unlike the video augmentation baseline, this experiment varies the amount of real expert behavior available for training.
On the contact-rich \texttt{Open Drawer} task, we extend the real-world demonstrations to 200 trajectories, vary the number used for training, and compare real-only SFT, SFT co-training, and \modify{RL-Co} in Fig.~\ref{fig:data_eef}.

As expected, all methods improve with additional real demonstrations, and SFT co-training benefits more rapidly than real-only training by using simulated data.
However, both supervised baselines remain substantially below or comparable to \modify{RL-Co} trained with far fewer real-world demonstrations, showing that \modify{RL-Co} can amplify limited real data through simulation interaction and provide a more data-efficient path to deployment than collecting additional physical demonstrations alone. \crv{Appendix~\ref{app:crv-cost} further reports resource estimates for simulator setup, data generation, and training.}

\section{Conclusion}

This paper proposes \modify{RL-Co}, an RL-based sim-real co-training framework for vision-language-action (VLA) models.
RL-Co addresses the limitation of supervised sim-real co-training by first initializing the policy with real and simulated demonstrations, then improving it through reinforcement learning in simulation while using real-world supervision to preserve deployable behaviors.

Experiments across \crv{four tabletop tasks and two VLA architectures} show that \modify{RL-Co} outperforms real-only fine-tuning and SFT-based co-training, improving real-world success, robustness to distribution shifts, and \crv{real-world data efficiency}.
\crv{Additional comparisons with SFT on RL-generated rollouts further support the benefit of improving the policy through online simulation interaction.}
These results suggest that interactive simulation can unlock benefits that static imitation objectives alone cannot fully realize.

\section{Limitations}
\crv{Our approach relies on constructing simulation environments, and the ManiSkill setup used in this work is limited to rigid-body simulation. Within this setting, we evaluate RL-Co on four tabletop tasks using a single robot embodiment, with two evaluation sessions per condition. More challenging tasks, complex environments, and different robot embodiments remain to be explored. Furthermore, the sim-to-real gap continues to limit real-world performance, and success rates remain below 100\% in several settings despite the improvements achieved by RL-Co. Future work will address these limitations by broadening the simulation and evaluation settings and improving sim-to-real transfer.}

\clearpage
\acknowledgments{This work was supported by National Natural Science Foundation of China (No.~62406159), Zhongguancun Academy (Grant No.~C20250301), and Shenzhen Science and Technology Program (No.~AI2026016).}

\bibliography{references}

\clearpage
\appendix
\section*{Appendix}

\captionsetup[table]{position=bottom,skip=7pt}

\section{Detailed Problem Setup and Post-Training Objectives}
\label{app:problem-formulation}

\subsection{Problem Formulation}

We consider a two-domain robotic manipulation setting with a real-world task domain $T_{\text{real}}$ and a corresponding simulation task domain $T_{\text{sim}}$. 
The simulation environment is designed to match the real-world setup at the task level while allowing scalable data collection through interaction.

We model both the real-world task and its simulated counterpart as Partially Observable Markov Decision Processes (POMDPs), 
denoted by the tuple
\begin{equation}
\mathcal{M}_{\Omega} = \langle 
\mathcal{S}_{\Omega}, 
\mathcal{A}, 
\mathcal{P}_{\Omega}, 
\mathcal{R}, 
\mathcal{O}_{\Omega}, 
\mathcal{L}, 
P(s_0), 
\gamma 
\rangle,
\end{equation}
where $\Omega \in \{\text{real}, \text{sim}\}$ indicates whether the process corresponds to the real-world or simulation task.

Following the formulation in~\citet{maddukuri2025sim}, we define each component as follows:

\begin{itemize}
    \item $\mathcal{S}_{\Omega}$ and $\mathcal{O}_{\Omega}$ denote the state space of the robot-environment system and the observation space induced by onboard sensors, respectively. 
    While the real and simulated tasks operate in different environments, they share the same robot embodiment and sensing modalities.

    \item $\mathcal{A}$ is the robot action space. 
    Both tasks adopt an identical control interface and action parameterization.

    \item $\mathcal{P}_{\Omega}$ represents the state transition dynamics, where $s_{t+1} \sim \mathcal{P}_{\Omega}(\cdot \mid s_t, a_t)$. 
    Due to the inherent difficulty of perfectly modeling real-world physics, the transition dynamics in simulation may exhibit slight discrepancies from those in the real environment.

    \item $\mathcal{L}$ denotes the natural language instruction specifying the task goal. 
    For corresponding real and simulated tasks, the language instruction remains identical.

    \item $\mathcal{R}$ is the reward function, defined as $\mathcal{R}(s, l)$, which evaluates task progress based on the current state and the given language instruction.

    \item $P(s_0)$ is the distribution over initial states, from which $s_0 \sim P(s_0)$ is sampled. 
    The real and simulated tasks share the same initial state distribution.

    \item $\gamma \in (0,1)$ is the discount factor.
\end{itemize}

Under this formulation, we define a vision-language-action (VLA) policy $\pi_{\theta}$ that conditions on the most recent $H$ observations $o_{\Omega}^{t-H+1:t}$ and the language instruction $l$ to predict a sequence of future actions over a horizon of length $h$:
\begin{equation}
a_{t:t+h-1} \sim \pi_{\theta}\bigl(a_{t:t+h-1} \mid o_{\Omega}^{t-H+1:t}, l\bigr).
\end{equation}

\subsection{Fine-Tuning on Vision-Language-Action Models}

We consider post-training of vision-language-action (VLA) models under both supervised and reinforcement learning paradigms. 
Given a pre-trained VLA policy $\pi_{\theta}$, fine-tuning aims to adapt the policy to a specific manipulation task by leveraging either expert demonstrations or online interaction with the environment.

\subsubsection{Supervised Fine-Tuning (SFT)}

Given an expert-collected demonstration dataset 
$\mathcal{D}_{T} = \{(\tau^{(i)}, l^{(i)})\}_{i=1}^{N}$, 
each trajectory $\tau^{(i)} = \{(o^{(i)}_j, a^{(i)}_j)\}_{j=1}^{K_i}$ consists of paired observations and actions, 
and $l^{(i)}$ denotes the corresponding natural language instruction. 
Here, $N$ is the total number of trajectories and $K_i$ is the length of the $i$-th trajectory.

Supervised fine-tuning optimizes the VLA policy $\pi_{\theta}$ by minimizing the discrepancy between predicted and expert actions:
\begin{equation}
\label{eq:sft_loss}
L_{\mathrm{SFT}}(\theta)
=
\underset{\substack{
(\tau, l) \sim D_T \\
t \sim \mathrm{Unif}(\{1,\dots,K_\tau\})
}}{\mathbb{E}}
\Big[
\ell_{\mathrm{SFT}}\!\left(
\hat{a}_{t:t+h-1},\,
a_{t:t+h-1}
\right)
\Big],
\end{equation}
where
\begin{equation}
\hat{a}^{(i)}_{t:t+h-1}
=
\pi_{\theta}\bigl(o^{(i)}_{t-H+1:t},\, l^{(i)}\bigr)
\end{equation}
denotes the predicted action chunk of horizon $h$, and
\begin{equation}
a^{(i)}_{t:t+h-1}
=
\{a^{(i)}_t, a^{(i)}_{t+1}, \dots, a^{(i)}_{t+h-1}\}
\end{equation}
is the corresponding expert action sequence.

The loss function $\ell_{\text{SFT}}$ depends on the specific VLA architecture and action representation. 
Common choices include next-token prediction losses~\cite{kim2024openvla}, 
$L_1$ regression losses for continuous actions~\cite{kim2025fine}, 
and diffusion-based denoising objectives~\cite{black2024pi_0}.

\subsubsection{Reinforcement Learning (RL) Fine-Tuning}

Reinforcement learning fine-tuning seeks to further optimize the policy through interaction with the environment by maximizing the expected discounted return:
\begin{equation}
\pi^{*}
=
\arg\max_{\pi_{\theta}}
\mathbb{E}_{\pi_{\theta}, \mathcal{P}}
\left[
\sum_{t=0}^{\infty}
\gamma^{t}
\mathcal{R}(s_t, l)
\right],
\end{equation}
where actions are sampled from the VLA policy $a_t \sim \pi_{\theta}(\cdot \mid o_t, l)$ and state transitions follow $s_{t+1} \sim \mathcal{P}(s_t, a_t)$.

Due to differences in action representations and generative mechanisms, the concrete realization of RL fine-tuning varies across VLA architectures.
Nevertheless, existing RL fine-tuning approaches share a common structure: 
an iterative loop of environment interaction for trajectory collection, followed by policy updates guided by reward feedback.
Our method builds upon this general framework and introduces an additional supervised fine-tuning objective on real-world data during the policy update phase, 
which is compatible with a wide range of RL fine-tuning strategies.

\section{Dataset Generation Details}
\label{app:dataset-generation}

\subsection{Real-World Demonstrations}

For all four tasks, we collect expert demonstrations via human teleoperation using a 3D SpaceMouse.
Each trajectory starts from the same initial conditions as evaluation: the robot is reset to a fixed configuration, while task-relevant objects are randomly placed on the table.
Expert actions are recorded as end-effector delta control commands.
For each task, we collect 20--50 successful trajectories, forming the real-world dataset $\mathcal{D}_{\text{real}}$.

\subsection{Simulation Dataset Generation}

To scale up training data in simulation, we adopt MimicGen~\cite{mandlekar2023mimicgen} to generate large numbers of successful trajectories.
Instead of collecting teleoperated demonstrations directly in simulation, we replay real-world expert trajectories in ManiSkill and use them as seed trajectories for data generation, thereby grounding the simulation data in real-world behaviors.

We implement the MimicGen pipeline within ManiSkill and introduce a minor modification: for each seed trajectory, we retain only task-relevant key stages and remove long segments of free-space end-effector motion.
This pruning encourages smoother and more efficient generated trajectories.
For each task, we generate 1{,}000 successful trajectories, which together form the simulation dataset $\mathcal{D}_{\text{sim}}$.

\section{Real-world Environment Setup}
\label{app:real-world-setup}

\begin{figure} [h]
    \centering
    \includegraphics[width=0.6\linewidth]{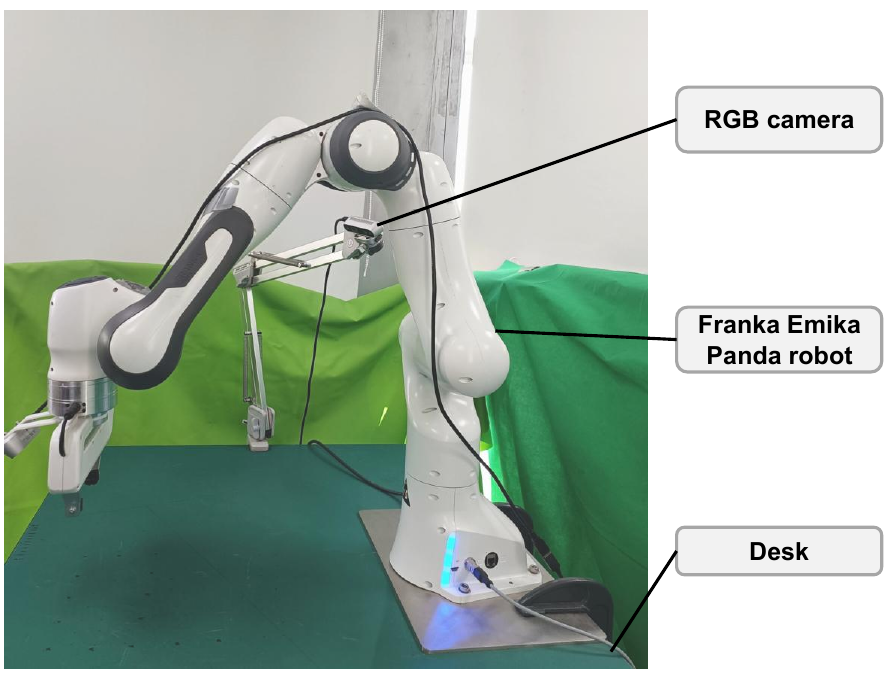}
    \caption{
    \textbf{Real-world Setup.} The real-world evaluation platform includes a tabletop workspace, a Franka Panda robotic manipulator fixed to the table, and a RGB camera for visual perception. All objects are positioned on the table surface.
    }
    \label{fig:real-setup}
\end{figure}

Fig.~\ref{fig:real-setup} illustrates our real-world evaluation setup.
The system consists of a table-top workspace, a Franka Emika Panda robot mounted on the table,
and a fixed RGB camera.
We use the RGB channels captured by the camera as the visual input to the VLA model.

The robot is equipped with seven actuated joints and a parallel-jaw gripper with open/close capability.
Control is performed in an end-effector delta-pose space:
at each timestep, we command a relative end-effector pose with respect to the current pose,
and compute the corresponding joint updates using inverse kinematics (IK).
The translational components are specified in the robot base frame,
while the rotational components are represented as roll--pitch--yaw (RPY) angles
relative to the current end-effector orientation.
Gripper actuation is controlled separately using a binary open/close signal.
Overall, the action space is 7-dimensional.

\section{Evaluation Details}
\label{app:evaluation-details}

\subsection{\crv{Evaluation Sessions and Statistical Interpretation}}
\label{app:crv-evaluation}

\crv{Here we present our evaluation protocol. For each checkpoint, we conduct two separate evaluation sessions. Each session contains 32 trials for \texttt{Pick and Place}, 30 for \texttt{Push Cube}, 10 for \texttt{Open Drawer}, and 10 for \texttt{Close Drawer}, giving 64, 60, 20, and 20 trials in total, respectively.}

\crv{Let $k_{t,s}$ be the number of successes for task $t$ in session $s\in\{1,2\}$, and let $n_t$ be the number of trials in that session. Define $r_{t,s}=100k_{t,s}/n_t$. The reported task mean and population standard deviation are}
\crv{\begin{equation}
\bar r_t=\frac{r_{t,1}+r_{t,2}}{2},\qquad
\sigma_t=\sqrt{\frac{(r_{t,1}-\bar r_t)^2+(r_{t,2}-\bar r_t)^2}{2}}.
\end{equation}}
\crv{For example, the OpenVLA real-only policy succeeds in $2/32$ \texttt{Pick and Place} trials in each session, yielding $6.25\%\pm0\%$ before rounding. The zero standard deviation reflects identical success rates in the two sessions. For the average across the four tasks, we first compute $a_s=\tfrac14\sum_{t=1}^4 r_{t,s}$ and then apply the same mean and standard deviation calculation across the two $a_s$. Thus tasks have equal weight despite different trial counts. Aggregates are computed before rounding.}

\crv{The standard deviations summarize session-level evaluation variability for the selected checkpoints. Estimating training-seed variance and establishing per-task confidence require additional independently trained policies and larger evaluations. Trial-level pooling and paired comparisons depend on the independence and matching of the recorded initial conditions. These considerations define the statistical scope of the current results.}

\subsection{Visualization of All Tasks}

\begin{figure*}
    \centering
    \includegraphics[width=0.9\linewidth]{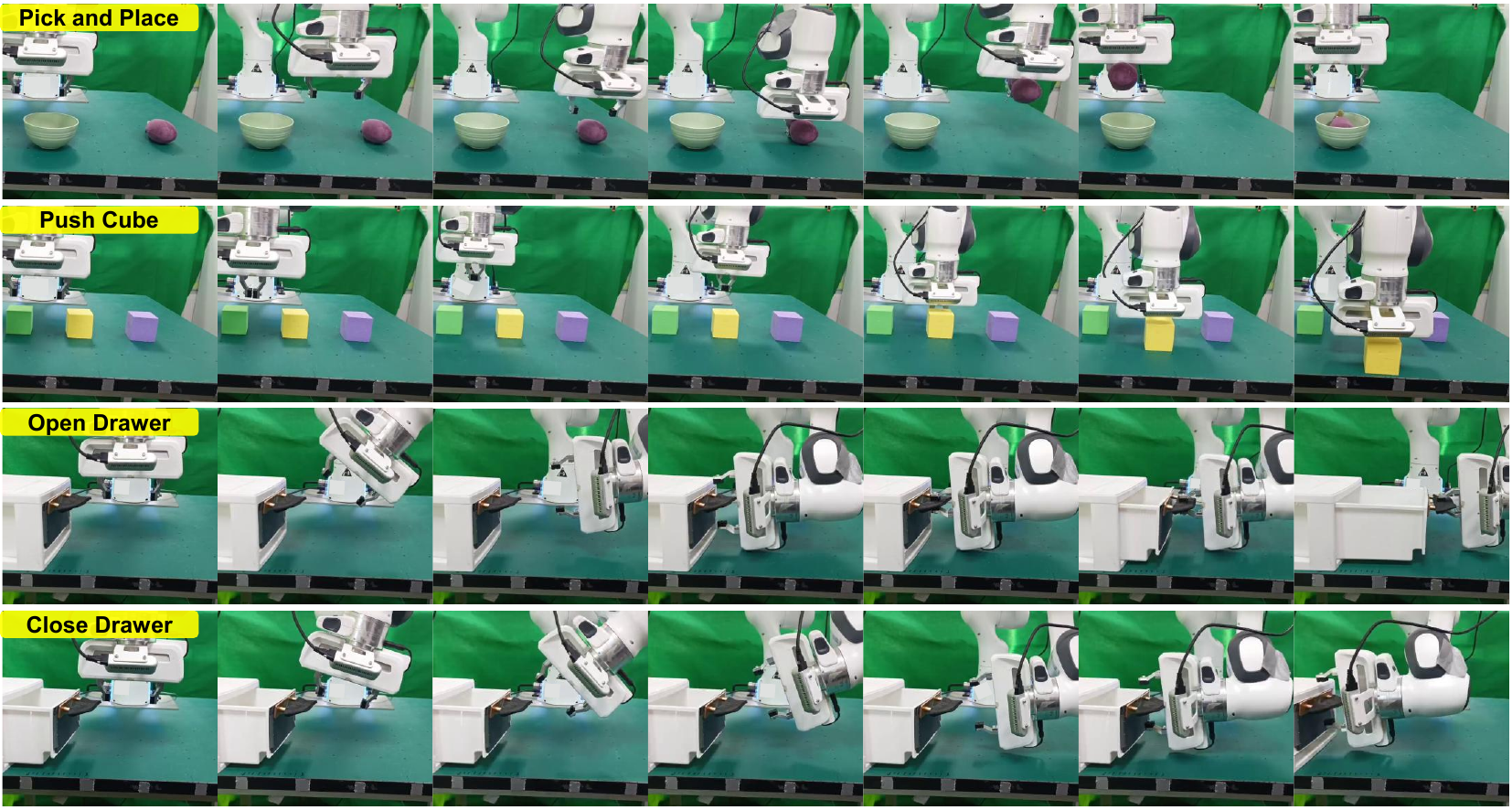}
    \caption{
    \textbf{Visualization of Four Tabletop Manipulation Tasks.}
    For each task, we present one successful trajectory and uniformly sample seven frames along the execution.
    Each row corresponds to a single trajectory shown from start to completion.
    }

    \label{fig:task-show}
\end{figure*}

Fig.~\ref{fig:task-show} visualizes the four table-top manipulation tasks evaluated in our experiments.
\texttt{Pick and Place} requires the robot to grasp an object from the table and place it into a bowl.
\texttt{Push Cube} involves pushing one cube out of three candidates with different colors according to a language instruction.
\texttt{Open Drawer} requires opening a closed drawer placed on the table,
while \texttt{Close Drawer} requires closing an initially opened drawer.

\subsection{Manipulated Objects}

\begin{figure}
    \centering
    \includegraphics[width=0.75\linewidth]{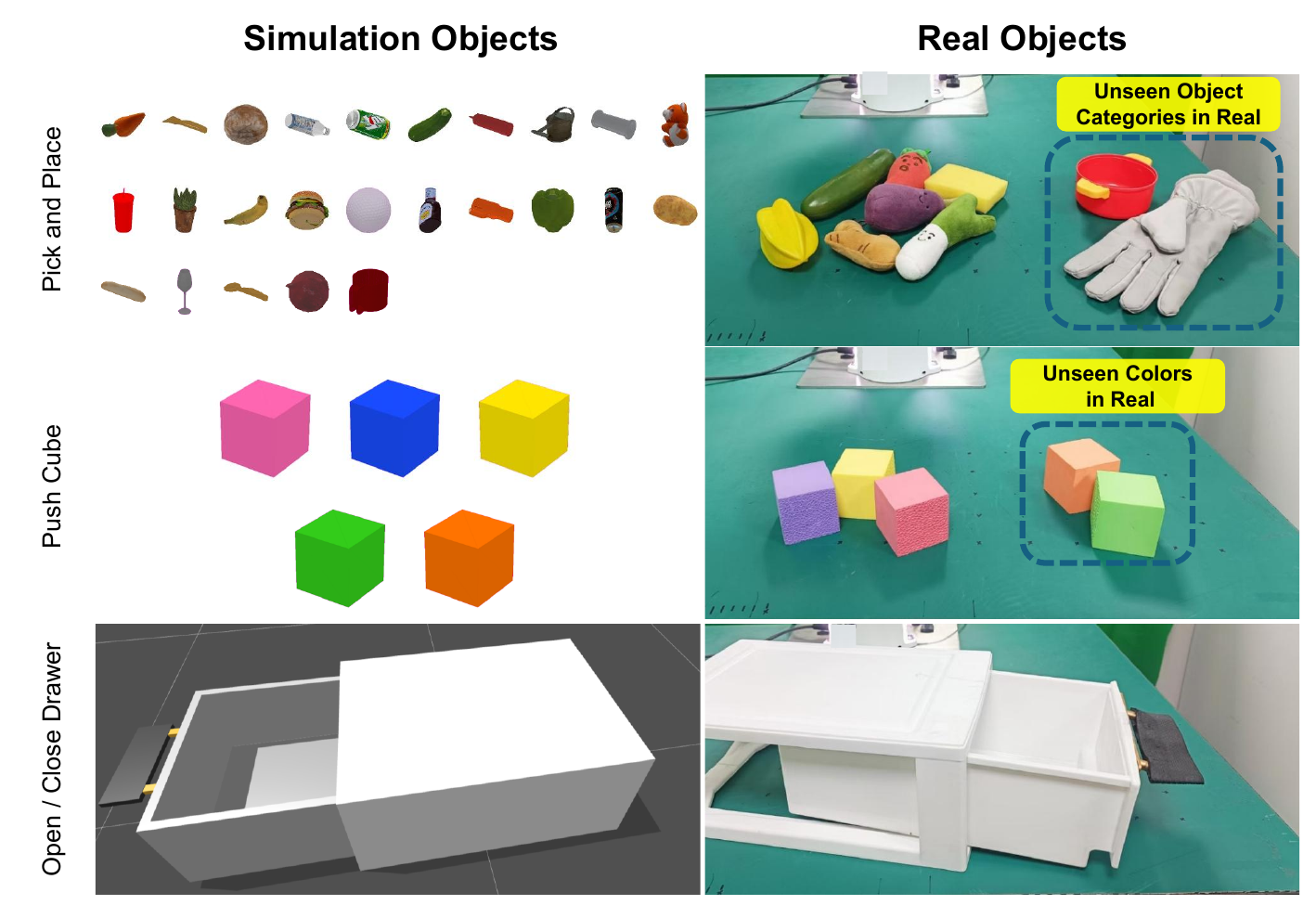}
    \caption{
    \textbf{Manipulated Objects in Simulation and the Real World.}
    The left panel shows the objects used in simulation, while the right panel presents the real-world objects.
    All simulated objects are used during training.
    The real-world objects are divided into training objects and unseen objects for generalization evaluation.
    }

    \label{fig:gen}
\end{figure}

Fig.~\ref{fig:gen} shows the manipulated objects used in both simulation and real-world experiments.
The detailed settings are summarized below:

\begin{itemize}
    \item \texttt{Pick and Place}:
    In simulation, we use the same set of 25 objects as in the environment proposed by~\citet{liu2025can}.
    In the real world, objects are divided into two categories: regular-shaped and irregular-shaped.
    Regular-shaped objects consist of toy fruits and vegetables,
    while irregular-shaped objects include bowls and gloves.
    Notably, irregular-shaped objects are not included in the real-world expert demonstrations.
    For in-distribution evaluation, we select four regular-shaped objects for testing.

    \item \texttt{Push Cube}:
    In simulation, we train on five colored cubes, as shown in Fig.~\ref{fig:gen}.
    In the real-world setup, we also use five colors.
    However, expert demonstrations are collected only for three colors (purple, yellow, and pink),
    while orange and green cubes are excluded from the demonstration data.
    During evaluation, three colors are randomly selected from the five available colors.

    \item \texttt{Open/Close Drawer}:
    In the real world, we use the drawer shown in Fig.~\ref{fig:gen}.
    In simulation, we construct a corresponding URDF model with matched geometric proportions.
\end{itemize}

\subsection{Objects Initial States}

\begin{figure}
    \centering
    \includegraphics[width=0.75\linewidth]{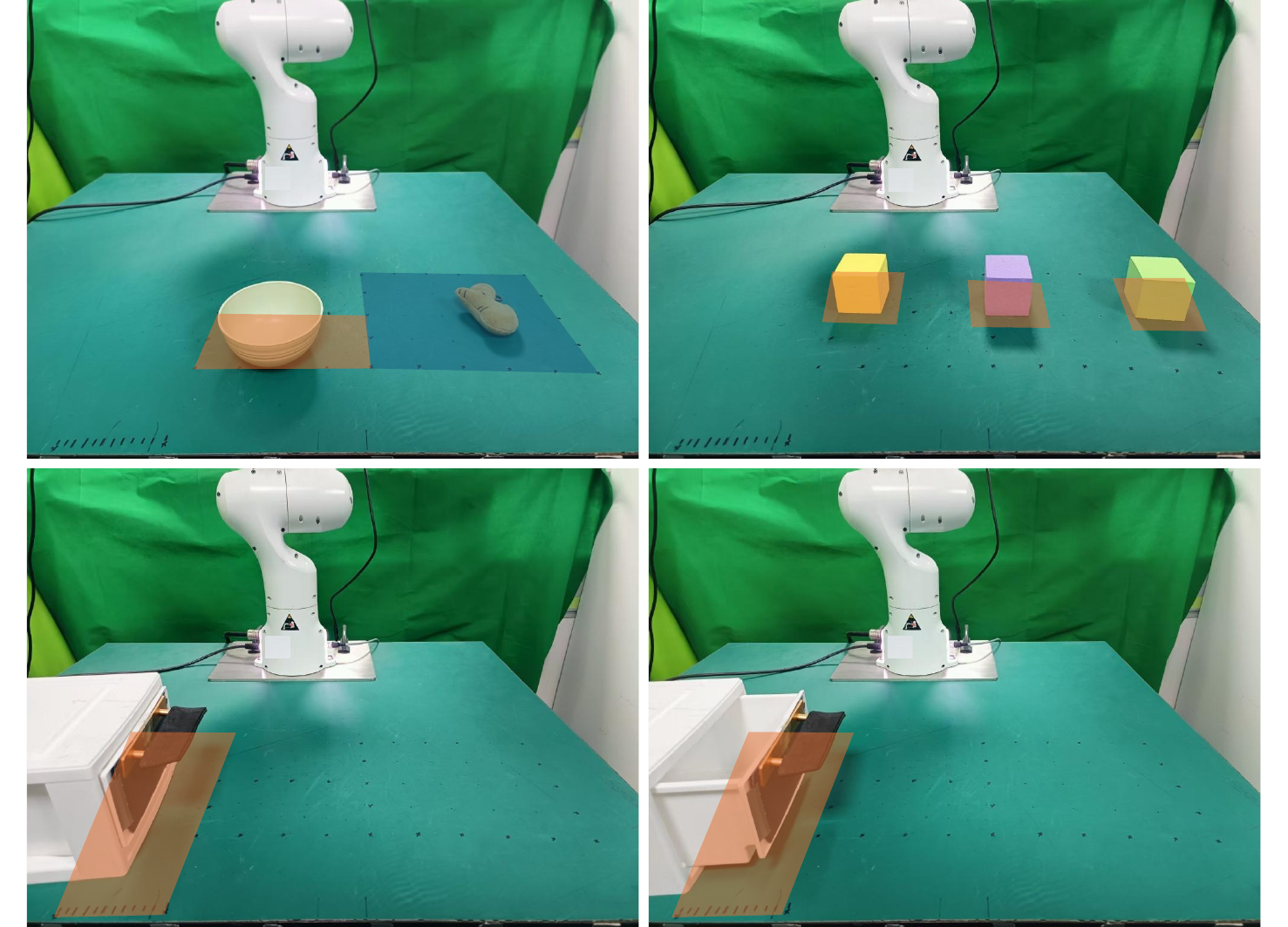}
    \caption{
    \textbf{Initial Regions for Manipulative Objects.}
    For the \texttt{Pick and Place} task, the bowl is placed within the orange region, while the objects are initialized in the blue region.
    For the \texttt{Push Cube} task, each cube is initialized within its corresponding orange region.
    For the \texttt{Open / Close Drawer} tasks, the front edge of the drawer is initialized within the orange region.
    }

    \label{fig:pos}
\end{figure}

Fig.~\ref{fig:pos} illustrates the randomized regions for the four real-world tasks.
The detailed configurations are as follows:

\begin{itemize}
    \item \texttt{Pick and Place}:
    The bowl is randomly placed within a $10 \times 20$\,cm rectangular region, indicated by the orange area in Fig.~\ref{fig:pos}.
    For each episode, one object is selected from a predefined object set,
    and its center is randomly placed within a $20 \times 25$\,cm rectangular region, indicated by the blue area in Fig.~\ref{fig:pos}.
    To facilitate controlled evaluation, both the bowl and object regions are discretized into grids with a minimum resolution of $5$\,cm.
    All objects are placed on grid points, and the same set of initial configurations is used across different methods.

    \item \texttt{Push Cube}:
    For each evaluation episode, three cubes are randomly selected from all available colors and randomly ordered.
    The cubes are initially placed with a spacing of $15$\,cm, followed by a random perturbation within a $5 \times 5$\,cm region,
    as indicated by the orange area in Fig.~\ref{fig:pos}.
    The language instruction specifies one of the three colors.
    All experiments follow the same color permutations and spatial configurations.

    \item \texttt{Open Drawer}:
    The front edge of the closed drawer is placed within the orange region shown in Fig.~\ref{fig:pos},
    with the drawer orientation initially aligned parallel to the short edge of the table.
    A random rotational perturbation of up to $15^\circ$ is then applied.
    We uniformly sample 10 predefined initial configurations, which are shared across all evaluations.

    \item \texttt{Close Drawer}:
    Similar to \texttt{Open Drawer}, the drawer is initially opened by approximately 10\,cm,
    and its front edge is placed within the same orange region with up to $15^\circ$ of rotational perturbation.
    The same set of 10 predefined configurations is used for evaluation to ensure fair comparison.
\end{itemize}

\subsection{Robot Initial State}

Unless otherwise specified, the Franka Emika Panda robot is initialized in a fixed default configuration across all experiments, as shown in Fig.~\ref{fig:real-setup}.
Here we describe additional robot initial states used in the generalization experiments.
Specifically, we focus on the \texttt{Pick and Place} task and select four representative objects, each with a fixed object initialization.
For each object, we perturb the robot tool center point (TCP) by applying a rotation of $\pm 30^\circ$ around the vertical axis, together with a translational offset of 5\,cm along a single Cartesian direction.
The perturbations include forward, backward, leftward, rightward, and upward translations, resulting in five distinct perturbed initial states.
Each perturbation combines one directional translation with the corresponding rotational offset.
These perturbations are summarized in Table~\ref{tab:initial_state_perturbation}.
All other aspects of the environment and policy remain unchanged.

\begin{table}[t]
    \centering
    \begin{tabular}{c c c}
    \toprule
    \textbf{Perturbation ID} & \textbf{Translation (cm)} & \textbf{Rotation (deg)} \\
    \midrule
    P1 & $(+5,\,0,\,0)$  (forward)   & $+30^\circ$ \\
    P2 & $(-5,\,0,\,0)$ (backward)  & $-30^\circ$ \\
    P3 & $(0,\,+5,\,0)$ (left)      & $+30^\circ$ \\
    P4 & $(0,\,-5,\,0)$ (right)     & $-30^\circ$ \\
    P5 & $(0,\,0,\,+5)$ (upward)    & $+30^\circ$ \\
    \bottomrule
    \end{tabular}
    \caption{\textbf{Robot initial state perturbations applied to the TCP in the \texttt{Pick and Place} task.}
    Translations are defined in the world frame, and rotations are applied around the vertical ($z$) axis.}
    \label{tab:initial_state_perturbation}
\end{table}

\subsection{Success Criteria and Failure Modes}

In all real-world evaluations, we use the same sampled initial configurations across methods and judge task success according to task completion rather than intermediate motion quality.
The task-specific criteria are:

\begin{itemize}
    \item \texttt{Pick and Place}: An episode is successful only if the robot grasps the target object and releases it inside the bowl. Failure cases include missing the grasp, dropping the object before reaching the bowl, or placing the object outside the bowl.
    \item \texttt{Push Cube}: An episode is successful if the robot pushes the cube specified by the language instruction forward. Pushing an incorrect cube or failing to make effective contact with the instructed cube is counted as failure.
    \item \texttt{Open Drawer}: An episode is successful if the robot grasps or makes stable contact with the drawer handle and pulls the drawer open beyond the task threshold. Common failures include missing the handle, losing contact during pulling, or pulling in a direction that does not open the drawer sufficiently.
    \item \texttt{Close Drawer}: An episode is successful if the robot pushes the opened drawer back to the closed state. Failures typically occur when the robot contacts an ineffective part of the drawer, pushes in an incorrect direction, or stops before the drawer is fully closed.
\end{itemize}

\section{Simulation Training}
\label{app:simulation-training}

\subsection{Simulation Fidelity and Sim--Real Alignment}

\textbf{Visual and task-level alignment.}
\crv{We first fixed the real robot, table, camera, and task workspace, and then built the simulator from approximate measurements of that setup.}
Our simulation environments are constructed in ManiSkill~\cite{tao2024maniskill3} to preserve the task-relevant structure of the corresponding real-world setups.
The simulated and real environments share the same robot embodiment, end-effector action interface, task semantics, object-level spatial layouts, and camera viewpoint used by the VLA policy.
We manually tune the simulated camera extrinsics so that the rendered third-person view approximately matches the fixed RGB camera used in real-world deployment.
We also align the table placement, robot mounting pose, task workspace, and high-level object arrangements so that the simulated rollouts expose the policy to interaction patterns that are meaningful for the real task.

We do not assume that the simulator is a photorealistic or fully accurate digital twin.
Object identities, materials, textures, lighting, and background appearance are not precisely matched between simulation and the real world.
This choice is intentional: our goal is not to demonstrate zero-shot transfer from a high-fidelity simulator, but to test whether imperfect yet task-relevant simulation can provide useful closed-loop interaction when the policy is anchored by real-world demonstrations.
The visual gap is therefore handled by the real-data supervision in both stages of \modify{RL-Co}, rather than by requiring exact visual reconstruction.

\textbf{Physics abstraction.}
For physics, we use rigid-body simulation in ManiSkill.
Simulated objects are imported from mesh files, and their collision geometries are approximated by convex hulls computed with CoACD~\cite{wei2022approximate}.
This abstraction is sufficient for representing the dominant contacts in our tabletop tasks, but it does not fully capture all real-world effects such as soft-object deformation, small unmodeled contacts, or friction variations.
Thus, we view the simulator as a task-level interaction environment rather than a complete physical replica.

\textbf{Robot controller system identification.}
To better align the low-level robot response, we perform a lightweight system identification step for the Franka controller.
We first collect one teleoperated real-world trajectory containing the commanded 7-DoF actions and the corresponding tool-center-point (TCP) trajectory.
We then replay the same action sequence in simulation and optimize the simulated Franka joint stiffness and damping parameters using simulated annealing.
After tuning, the simulated TCP trajectory stays within approximately 2\,cm of the real TCP trajectory over the replayed sequence.
All other simulator parameters remain at their default ManiSkill values.

\crv{These choices follow the practical importance of viewpoint alignment in prior sim-real co-training~\cite{maddukuri2025sim} and of physical correspondence in planar pushing~\cite{wei2025empirical}. Approximate task geometry and camera/controller alignment define our evaluated setup. The approximately 2\,cm TCP error characterizes controller replay on the calibration trajectory. Systematic perturbations of geometry, camera pose, and controller parameters would further characterize the transfer tolerance of each component.}

\subsection{Reward Function Design}

We detail the reward function design for each simulation task in this section.

\begin{itemize}
    \item \texttt{Pick and Place}:
    This task is decomposed into two sequential stages: \emph{grasping} and \emph{placing}, which are indicated by the binary states \texttt{is\_grasped} and \texttt{is\_placed}, respectively. We design two types of reward functions: a dense reward and a sparse reward. The dense reward is defined as
    \begin{equation}
        \mathcal{R} = \min\Bigl\{\mathbb{I}_{\text{success}},\,
        \bigl(\mathbb{I}_{\text{grasped}}(1+\mathcal{R}_d(d_2))+\mathcal{R}_d(d_2)\bigr)\Bigr\},
    \end{equation}
    where $\mathbb{I}_{\text{grasped}}$ and $\mathbb{I}_{\text{success}}$ are indicator functions denoting whether the object has been successfully grasped and placed, respectively.
    The shaping term $\mathcal{R}_d(x)=1-\tanh(10x)$ provides a smooth distance-based reward that asymptotically approaches 1 as the distance decreases.
    Here, $d_1$ and $d_2$ denote the distance between the gripper and the object, and the distance between the object and the target container, respectively.
    For the sparse reward, we assign a reward of $0.2$ at the moment when grasping succeeds, and a reward of $1$ upon successful placement. If the object leaves the target container after a successful placement due to external disturbances, a penalty of $-0.4$ is applied. All other timesteps receive zero reward.
    We use dense reward when training OpenVLA and use sparse reward when training $\pi_{0.5}$.

    \item \texttt{Push Cube}:
The objective of this task is to push a designated target cube into a predefined goal region on the table.
The dense reward consists of three components.
First, a \emph{reaching reward} encourages the Tool Center Point (TCP) to approach a pre-defined pushing pose behind the cube along the pushing direction:
\begin{equation}
r_{\text{reach}} = 1 - \tanh\!\left(5 \cdot \lVert \mathbf{p}_{\text{tcp}} - \mathbf{p}_{\text{push}} \rVert_2 \right),
\end{equation}
where $\mathbf{p}_{\text{push}}$ is defined as a point offset from the cube center by one half cube length plus a small margin along the pushing axis.
Second, once the TCP is sufficiently close to the pushing pose, a \emph{placement reward} is activated to encourage the cube to move toward the goal region:
\begin{equation}
r_{\text{place}} = 1 - \tanh\!\left(5 \cdot \lVert \mathbf{p}_{\text{cube}}^{xy} - \mathbf{p}_{\text{goal}}^{xy} \rVert_2 \right),
\end{equation}
where only planar $(x,y)$ distances are considered.
This term is gated by a proximity condition to ensure that the agent first establishes contact before being rewarded for object motion.
Finally, a sparse \emph{success bonus} is assigned once the cube is pushed beyond the goal center along the pushing direction and remains within a tolerance band orthogonal to it:
\begin{equation}
r =
\begin{cases}
3.0, & \text{if success}, \\
r_{\text{reach}} + r_{\text{place}}, & \text{otherwise}.
\end{cases}
\end{equation}

\item \texttt{Open Drawer}:
In this task, the reward is defined over three stages corresponding to reaching, opening progress, and task completion.
First, a reaching reward encourages the TCP to approach the drawer handle:
\begin{equation}
r_{\text{reach}} = 1 - \tanh\!\left(5 \cdot \lVert \mathbf{p}_{\text{tcp}} - \mathbf{p}_{\text{handle}} \rVert_2 \right).
\end{equation}
Second, an \emph{opening reward} is defined based on the normalized drawer joint position (open fraction):
\begin{equation}
r_{\text{open}} = 2 \cdot \text{open\_frac},
\end{equation}
where the open fraction is computed by linearly normalizing the drawer joint position between its minimum and maximum limits.
Once the drawer begins to open, the reaching reward is saturated to a constant value to avoid conflicting gradients.
The total dense reward is given by:
\begin{equation}
r = r_{\text{reach}} + r_{\text{open}}.
\end{equation}
A terminal success reward of $5.0$ is assigned when the drawer is opened beyond a high open-fraction threshold (e.g., $90\%$ of its range).

\item \texttt{Close Drawer}:
The Close Drawer task is initialized from an open state and rewards progress toward closing the drawer.
Instead of absolute position, we define a \emph{progress-based reward} using the change in open fraction between consecutive time steps:
\begin{equation}
\Delta = \text{open\_frac}_{t-1} - \text{open\_frac}_{t}.
\end{equation}
The dense reward is then computed as:
\begin{equation}
r = \alpha \cdot \text{clip}(\Delta, -1, 1) - \beta,
\end{equation}
where $\alpha$ is a scaling factor for closing progress and $\beta$ is a small time penalty that encourages faster completion.
Once the drawer is closed below a predefined threshold on the open fraction, a terminal success reward of $5.0$ is issued and overrides the dense shaping terms.
\end{itemize}

For all tasks, dense rewards are normalized by their respective maximum achievable reward values to ensure comparable reward scales across tasks during multi-task training.

\subsection{Performance in Simulation during RL Training}
\label{app:sim-training}

\crv{Interaction budgets are fixed within each model--task configuration but differ across configurations. Task horizons, reward sparsity, exploration difficulty, and parallel rollout configurations all affect the number of transitions and elapsed time. Appendix~\ref{app:crv-cost} reports elapsed-time curves for the four $\pi_{0.5}$ runs and their hardware allocations.}

\begin{figure*}
    \centering
    \includegraphics[width=0.9\linewidth]{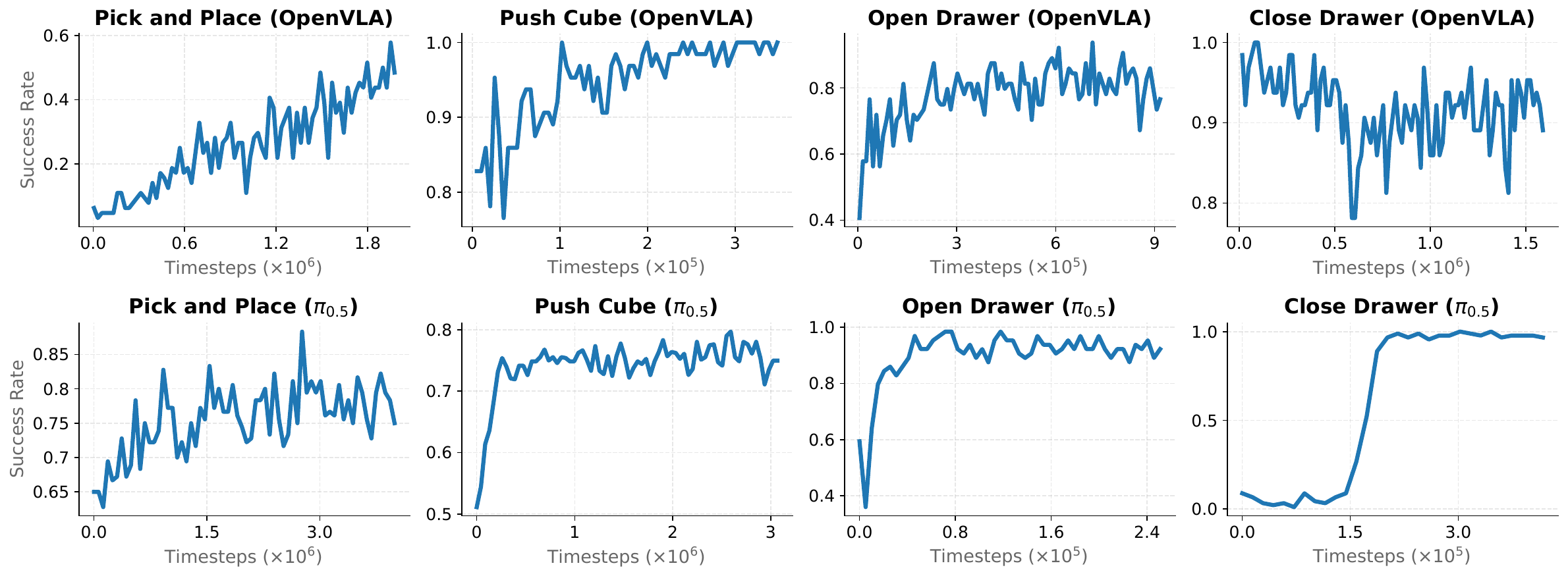}
    \caption{
    \textbf{Simulation Training Results.}
    We report the simulation success rates across all settings during RL training.
    }
    \label{fig:rl-curve}
\end{figure*}

Fig.~\ref{fig:rl-curve} shows the success rates of different models on each task in the simulation environment during RL training.
After RL fine-tuning, all VLA models achieve substantial performance improvements in simulation
compared to their pre-RL counterparts.

\section{Additional Dexterous Evaluation}
\label{sec:appendix_dexterous}

\begin{figure}[h]
    \centering
    \includegraphics[width=\linewidth]{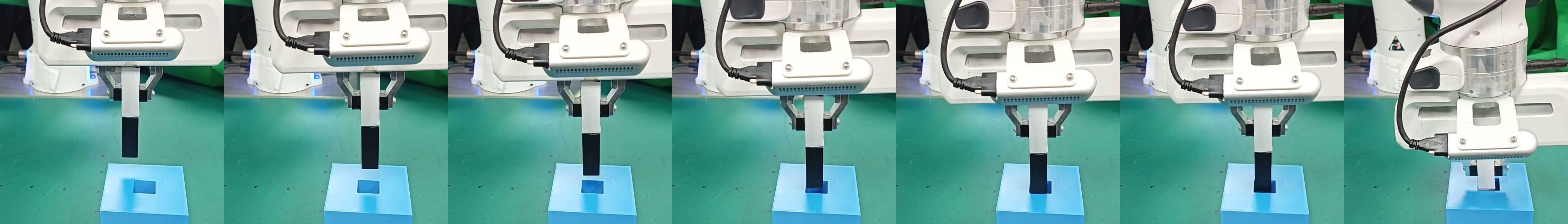}
    \caption{
    \textbf{Execution sequence of the \texttt{Peg Insertion} task.}
    The robot must align the grasped peg with the square hole and insert it into the base.
    }
    \label{fig:peg_insertion_appendix}
\end{figure}

We further evaluate \modify{RL-Co} on a more dexterous \texttt{Peg Insertion} task, which requires more precise contact and alignment than the tabletop tasks in the main experiments.
As shown in Fig.~\ref{fig:peg_insertion_appendix}, the robot starts by holding a $2 \times 2 \times 12$\,cm peg above a base with a $2.25 \times 2.25$\,cm square hole.
At the beginning of each episode, the peg is initialized 1--5\,cm above the hole, while the hole position is randomly perturbed in the horizontal plane.
An episode is successful only if the robot inserts the peg into the hole.

For this task, we collect 50 real-world demonstrations using human teleoperation and generate 500 simulated trajectories with motion planning.
We compare SFT-based sim--real co-training with \modify{RL-Co} using the $\pi_{0.5}$ backbone and the same real-world evaluation protocol.
As shown in Table~\ref{tab:peg_insertion_appendix}, \modify{RL-Co} improves real-world success from $32.5\%$ to $42.5\%$, suggesting that closed-loop simulation interaction can still provide additional gains in a more precision-demanding setting.

\begin{table}[h]
    \centering
    \small
    \begin{tabular}{l c}
    \toprule
        Method & Real SR (\%) \\
    \midrule
        SFT-based sim--real co-training & 32.5 \\
        \textbf{\modify{RL-Co} (Ours)} & \textbf{42.5} \\
    \bottomrule
    \end{tabular}
    \caption{
    \textbf{Real-world success rate on the \texttt{Peg Insertion} task.}
    The task uses the $\pi_{0.5}$ backbone, 50 real-world demonstrations, and 500 simulated trajectories.
    }
    \label{tab:peg_insertion_appendix}
\end{table}

\section{Sim--Real Correlation During RL Co-Training}
\label{sec:appendix_sim_real_correlation}

We additionally examine whether simulation performance provides a useful signal for real-world deployment during RL co-training.
We focus on the $\pi_{0.5}$ model on the \texttt{Pick and Place} and \texttt{Open Drawer} tasks.
During Stage~II RL co-training, we record the simulation success rate throughout training and periodically evaluate intermediate checkpoints on the real robot.
This analysis is intended to test whether the improvement observed in simulation corresponds to real-world improvement, rather than only reflecting simulator-specific artifacts.

\begin{figure}[h]
    \centering
    \includegraphics[width=0.75\linewidth]{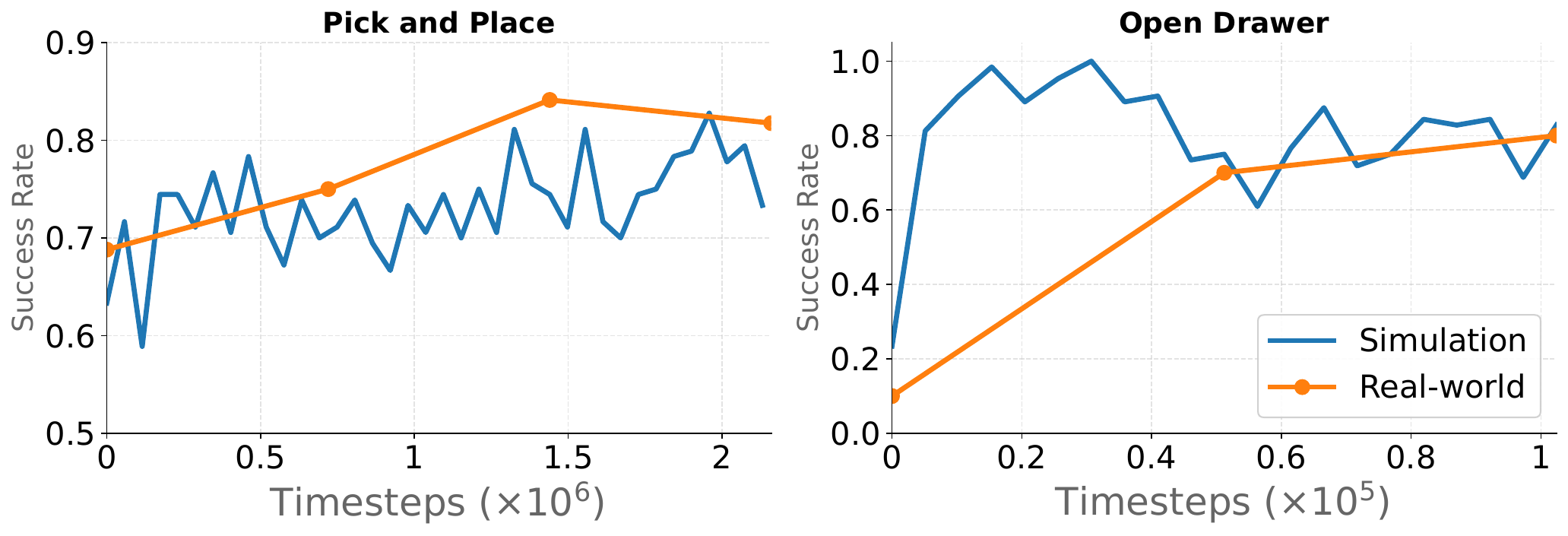}
    \caption{
    \textbf{Correlation between simulation and real-world success during RL co-training.}
    We evaluate intermediate checkpoints of $\pi_{0.5}$ on the \texttt{Pick and Place} and \texttt{Open Drawer} tasks.
    Real-world performance generally improves together with simulation performance.
    }
    \label{fig:sim_real_correlation_appendix}
\end{figure}

As shown in Fig.~\ref{fig:sim_real_correlation_appendix}, the real-world success rate generally increases as the simulation success rate improves.
This positive correlation supports our checkpoint-selection strategy based on simulation success convergence.
At the same time, the correlation is not perfect, which is expected under imperfect sim--real alignment.
We therefore use simulation convergence as a practical selection signal, while relying on real-world demonstrations in \modify{RL-Co} to reduce domain drift during policy improvement.

\section{Additional Baselines and Ablations}
\label{sec:appendix_additional_baselines}

We include additional comparisons on $\pi_{0.5}$ for the \texttt{Pick and Place} task to evaluate alternative ways of using simulation data.
These comparisons are not intended as a full benchmark suite, but as targeted tests of two natural alternatives to \modify{RL-Co}: removing the two-stage initialization, and distilling a simulation expert into the VLA policy through supervised trajectories.

\subsection{One-Stage RL Co-Training}

The first alternative removes Stage~I task-specific SFT initialization and directly applies Stage~II RL co-training to the pretrained VLA policy.
During this one-stage variant, the supervised regularization term is computed on the mixture of real and simulated demonstrations, rather than only on real demonstrations.
This design tests whether the SFT warm-up stage can be replaced by a single RL phase with an all-data SFT loss.

\begin{figure}[h]
    \centering
    \includegraphics[width=0.5\linewidth]{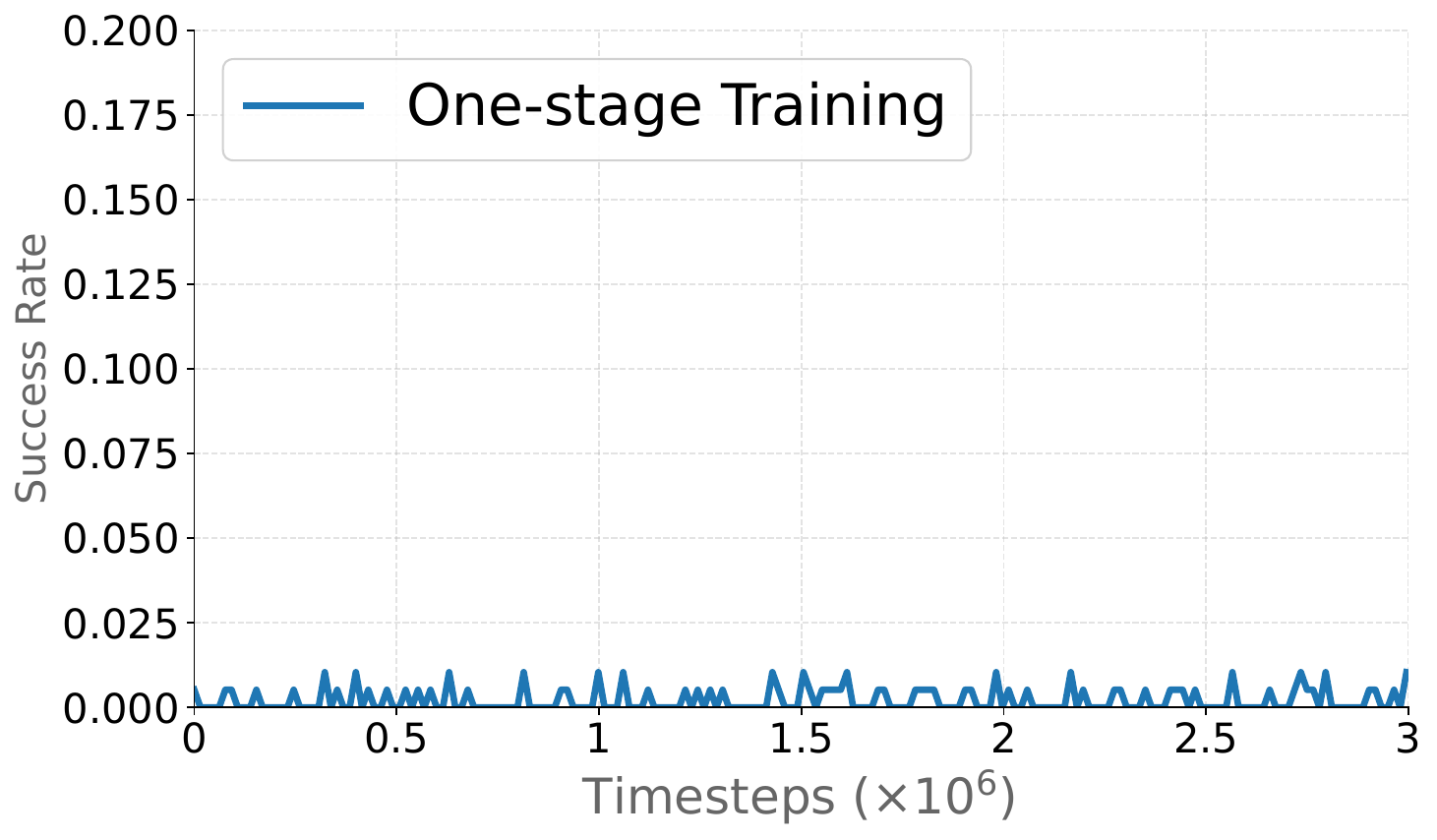}
    \caption{
    \textbf{Simulation success rate of one-stage RL co-training.}
    Without Stage~I task-specific SFT initialization, the policy maintains near-zero success in simulation even after several million interaction steps.
    }
    \label{fig:one_stage_appendix}
\end{figure}

The one-stage variant fails to acquire the task, achieving $0.0\%$ real-world success and maintaining near-zero simulation success even after several million interaction steps, as shown in Fig.~\ref{fig:one_stage_appendix}.
This result indicates that the SFT initialization stage is not merely a convenience for faster training; it provides the task-level behavior prior required for subsequent RL optimization.
This observation is consistent with the ablation in Section~\ref{sec:ablation-study}, where weakening the Stage~I initialization also makes RL substantially less effective.

\subsection{RialTo-Style Distillation}

The second alternative follows the distillation motivation of RialTo-style real-to-sim-to-real pipelines~\cite{torne2024reconcilingrealitysimulationrealtosimtoreal}.
We first train a simulation expert with RL, use the expert to generate 2{,}000 simulated trajectories, and then perform SFT co-training with these trajectories and the same 50 real-world demonstrations.
This baseline evaluates whether simulation is better used as an expert trajectory generator rather than as an interactive environment for directly improving the VLA policy.

The RialTo-style distillation baseline reaches $75.0\%$ real-world success, outperforming standard SFT co-training but remaining below \modify{RL-Co}.
This result suggests that a strong simulation expert can provide useful supervised trajectories, but distilling the expert into the VLA policy through static trajectories does not fully replace direct closed-loop policy improvement.
In contrast, \modify{RL-Co} updates the VLA policy itself through reward-driven interaction while using real demonstrations to preserve deployable real-world behavior.

\begin{table}[h]
    \centering
    \small
    \begin{tabular}{l c}
    \toprule
        Method & Real SR (\%) \\
    \midrule
        One-stage RL co-training & 0.0 \\
        RialTo-style distillation~\cite{torne2024reconcilingrealitysimulationrealtosimtoreal} & 75.0 \\
        \textbf{\modify{RL-Co} (Ours)} & \textbf{81.3} \\
    \bottomrule
    \end{tabular}
    \caption{
    \textbf{Additional baselines on \texttt{Pick and Place} with $\pi_{0.5}$.}
    Results are reported as real-world success rate without standard deviation because each additional baseline was evaluated once.
    }
    \label{tab:additional_baselines_appendix}
\end{table}

\subsection{\crv{Strong-Rollout SFT Control on Open Drawer}}
\label{app:crv-rollout}

\crv{We extend the rollout-distillation comparison to \texttt{Open Drawer} with $\pi_{0.5}$, where the main-table improvement over SFT co-training is largest. The additional SFT baseline combines the same 20 real demonstrations with $1.6\times10^5$ simulation transitions collected by an RL policy. This approximately matches RL-Co's combined MimicGen-generation and Stage~II interaction budget. Table~\ref{tab:crv-rollout} shows that the stronger simulated data improve SFT from 10\% to 20\%, while RL-Co reaches 65\%.}

\begin{table}[htbp]
\centering
\begin{tabular}{lc}
\toprule
\crv{Training condition ($\pi_{0.5}$, \texttt{Open Drawer})} & \crv{Real SR (\%)}\\
\midrule
\crv{SFT co-training with MimicGen demonstrations} & \crv{10.0}\\
\crv{SFT co-training with strong RL rollouts} & \crv{20.0}\\
\crv{RL-Co} & \crv{65.0}\\
\bottomrule
\end{tabular}
\caption{\crv{Strong-rollout SFT control using the same 20 real demonstrations.}}
\label{tab:crv-rollout}
\end{table}

\crv{This control reduces the confound between simulation-data quality and the training objective. The SFT baseline learns from a fixed rollout dataset, while RL-Co collects rollouts online as its policy changes. The budget match is defined in simulation transitions. Together with the \texttt{Pick and Place} distillation result, this comparison provides additional evidence for the benefit of direct RL updates in the tested settings.}

\section{Implementation Details}

\begin{table*}
    \centering
    \resizebox{\textwidth}{!}{%
    \begin{tabular}{c|c|cccc}
    \toprule
        Parameter Names
        & Setting
        & \texttt{Pick and Place}
        & \texttt{Push Cube}
        & \texttt{Open Drawer}
        & \texttt{Close Drawer} \\
    \midrule
        \multirow{1}{*}{General}
        & Number of Real Demos
        & 50 & 50 & 20 & 30 \\
    \midrule
        \multirow{5}{*}{OpenVLA}
        & Co-training ratio $\alpha$
        & 0.5 & 0.5 & 0.5 & 0.5 \\
        & SFT learning rate
        & $5\times 10^{-4}$ & $5\times 10^{-4}$ & $5\times 10^{-4}$ & $5\times 10^{-4}$ \\
        & Regularization weight $\beta$
        & 0.1 & 0.01 & 0.01 & 0.01 \\
        & Actor learning rate
        & $10^{-4}$ & $10^{-4}$ & $10^{-4}$ & $10^{-4}$ \\
        & Critic learning rate
        & $3\times 10^{-3}$ & $3\times 10^{-3}$ & $3\times 10^{-3}$ & $3\times 10^{-3}$ \\
    \midrule
        \multirow{6}{*}{$\pi_{0.5}$}
        & Co-Training Ratio $\alpha$
        & 0.5 & 0.95 & 0.95 & 0.98 \\
        & SFT learning rate
        & \crv{$2.5\times 10^{-5}$} & \crv{$2.5\times 10^{-5}$} & \crv{$2.5\times 10^{-5}$} & \crv{$2.5\times 10^{-5}$} \\
        & SFT lr schedule
        & Cosine Decay & Cosine Decay & Cosine Decay & Cosine Decay \\
        & Regularization Weight $\beta$
        & 1.0 & 0.2 & 1.0 & 0.1 \\
        & Actor learning rate
        & $4\times 10^{-6}$ & $10^{-6}$ & $4\times 10^{-6}$ & $7.91\times 10^{-6}$ \\
        & Critic learning rate
        & $2\times 10^{-4}$ & $2\times 10^{-4}$ & $2\times 10^{-4}$ & $1.55\times 10^{-4}$ \\
    \bottomrule
    \end{tabular}
    }
    \caption{
        \textbf{Hyperparameter settings for different VLA models and tasks.}
        We report task-specific hyperparameters used for OpenVLA and $\pi_{0.5}$ across four real-world manipulation tasks.
    }
    \label{tab:hyperparameter_settings}
\end{table*}

For OpenVLA, we fine-tune the model using LoRA with a rank of 32,
\crv{whereas $\pi_{0.5}$ updates only the action expert, with the vision-language backbone frozen.}
All models are optimized using the AdamW optimizer~\cite{loshchilov2017decoupled}.
The detailed hyperparameters are summarized in Table~\ref{tab:hyperparameter_settings}.

\crv{The real-data SFT anchor complements these parameter-update restrictions. Fig.~\ref{fig:ablation} measures its contribution by removing real supervision under the same training setup. Stage~II uses on-policy PPO~\cite{schulman2017proximal} with a learned value function, and the critic learning rates in Table~\ref{tab:hyperparameter_settings} govern PPO value updates.}

\section{\crv{Resource Accounting and Training Time}}
\label{app:crv-cost}

\subsection{\crv{Real-World Cost versus Simulation Cost}}

\crv{Fig.~\ref{fig:data_eef} compares the real-world demonstration requirements of the three training methods on \texttt{Open Drawer}. RL-Co achieves higher or comparable success with fewer real demonstrations, reducing the cost of real-world data collection. Beyond this real-world cost, the methods incur different simulation costs. Real-only SFT requires no simulation expenditure. SFT co-training additionally requires simulator construction and simulation-data generation, while RL-Co further incurs the cost of RL training in simulation.}

\crv{Simulation environments for different tasks can often share scene assets and setup effort. Simulation-data generation and training can also run unattended around the clock and be parallelized with additional computing resources, providing a low-cost way to scale experience collection. These advantages make simulation a practical substitute for additional human data collection, while its setup and computation costs remain part of the overall resource requirements. We therefore report a detailed time-cost comparison on \texttt{Open Drawer}, followed by training times across all four tasks.}

\Needspace{10\baselineskip}
\subsection{\crv{Time Cost on Open Drawer}}

\crv{Table~\ref{tab:crv-human-cost} summarizes the time required for simulator setup, real-world data collection, simulation-data generation, and RL co-training on \texttt{Open Drawer}. Constructing the simulator takes approximately one human working day, and this setup is shared across our four tasks. Collecting 200 real demonstrations takes approximately two human working days, with collection limited to working hours. In contrast, simulation-data generation and RL training can continue unattended outside working hours.}

\begin{table}[htbp]
\centering
\begin{tabular}{lll}
\toprule
\crv{Operation} & \crv{Time} & \crv{Note}\\
\midrule
\crv{Simulator setup} & \crv{$\sim$1 human working day} & \crv{Shared across four tasks}\\
\crv{Collect 200 real demonstrations} & \crv{$\sim$2 human working days} & \crv{Working hours only}\\
\crv{MimicGen generation} & \crv{1.5 h (RTX 4090)} & \crv{Unattended computation}\\
\crv{RL co-training} & \crv{9.43 h ($2\times$H100)} & \crv{Unattended computation}\\
\bottomrule
\end{tabular}
\caption{\crv{Time costs for \texttt{Open Drawer}. Simulator setup is shared across the four tasks. Computation times are wall-clock durations on the listed hardware.}}
\label{tab:crv-human-cost}
\end{table}

\crv{With 20 real demonstrations, RL-Co achieves higher real-world success than real-only SFT with 200 demonstrations in Fig.~\ref{fig:data_eef}. Its additional simulation work consists of the shared setup and approximately 11 hours of unattended data generation and RL training on the hardware in Table~\ref{tab:crv-human-cost}. Reusing the setup across tasks and running computation around the clock reduces the human effort required for policy improvement. This comparison shows the practical advantage of RL-Co in reducing real-world collection costs while achieving stronger performance.}

\subsection{\crv{RL Training Time}}

\begin{figure}[htbp]
\centering
\includegraphics[width=\linewidth]{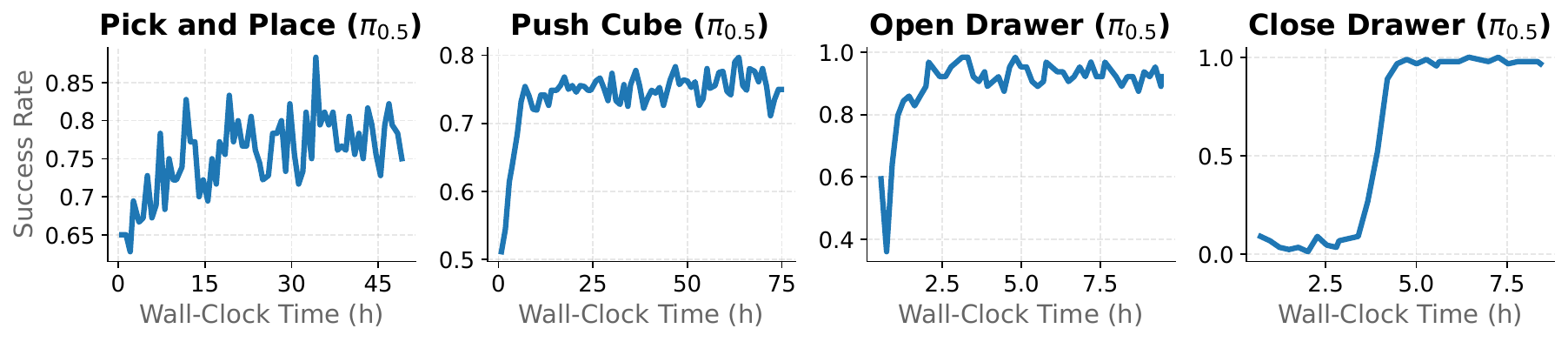}
\caption{\crv{Simulation success during $\pi_{0.5}$ RL co-training versus wall-clock time across four tasks.}}
\label{fig:crv-wallclock}
\end{figure}

\crv{Fig.~\ref{fig:crv-wallclock} reports simulation success against wall-clock time for the four $\pi_{0.5}$ Stage~II runs shown in Appendix~\ref{app:sim-training}. The runs use 3, 4, 2, and 3 H100 GPUs for \texttt{Pick and Place}, \texttt{Push Cube}, \texttt{Open Drawer}, and \texttt{Close Drawer}, respectively.}

\end{document}